\documentclass[preprint,12pt]{elsarticle}

\usepackage{amssymb}
\usepackage{amsmath}
\usepackage{times}
\usepackage{soul}
\usepackage{url}
\usepackage[hidelinks]{hyperref}
\usepackage[utf8]{inputenc}
\usepackage[small]{caption}
\usepackage{graphicx}
\usepackage{booktabs}
\usepackage{algorithmic}
\usepackage[switch]{lineno}
\usepackage{amssymb}
\usepackage[linesnumbered,ruled,vlined]{algorithm2e}
\usepackage{subfig}
\usepackage{float}
\usepackage[normalem]{ulem}
\usepackage{xcolor}
\usepackage{multirow} 
\usepackage{adjustbox}
\usepackage{graphicx}

\newtheorem{theorem}{Theorem}[section]

\newtheorem{pf}{Proof}

\newtheorem{definition}{Definition}[section]

\renewcommand{\qed}{\hfill\blacksquare}

\journal{Pattern Recognition}

\begin{document}

\begin{frontmatter}



\title{Nonlinear Performative Prediction}


\author[1]{Guangzheng Zhong} 
\ead{csgzzhong@comp.hkbu.edu.hk}

\author[1]{Yang Liu\corref{cor1}} 
\ead{csygliu@comp.hkbu.edu.hk}

\author[1]{Jiming Liu}
\ead{jiming@comp.hkbu.edu.hk}
\cortext[cor1]{Corresponding author}
\affiliation[1]{organization={Department of Computer Science, Hong Kong Baptist University},
            state={Hong Kong SAR},
            country={P.R. China}}

\begin{abstract}
Performative prediction is an emerging paradigm in machine learning that addresses scenarios where the model's prediction may induce a shift in the distribution of the data it aims to predict. Current works in this field often rely on uncontrollable assumptions, such as bounded gradients of performative loss, and primarily focus on linear cases in their examples and evaluations to maintain consistency between theoretical guarantees and empirical validations. However, such linearity rarely holds in real-world applications, where the data usually exhibit complex nonlinear characteristics. In this paper, we relax these out-of-control assumptions and present a novel design that generalizes performative prediction to nonlinear cases while preserving essential theoretical properties. Specifically, we formulate the loss function of performative prediction using a maximum margin approach and extend it to nonlinear spaces through kernel methods. To quantify the data distribution shift, we employ the discrepancy between prediction errors on these two distributions as an indicator, which characterizes the impact of the performative effect on specific learning tasks. By doing so, we can derive, for both linear and nonlinear cases, the conditions for performative stability, a critical and desirable property in performative contexts. Building on these theoretical insights, we develop an algorithm that guarantees the performative stability of the predictive model. We validate the effectiveness of our method through experiments on synthetic and real-world datasets with both linear and nonlinear data distributions, demonstrating superior performance compared to state-of-the-art baselines.
\end{abstract}



    
    

\begin{keyword}
Performative Prediction \sep Performative Stability \sep Data Distribution Map \sep Repeated Risk Minimization


\vspace{0.5cm}
\end{keyword}

\end{frontmatter}



\section{Introduction}
Performative prediction, introduced by~\cite{performative}, describes a scenario where the deployment of a prediction model causes a shift in the data distribution it aims to predict. This concept has been extensively investigated across various domains involving decision-making or human interaction, such as recommendation system \cite{performativerecommendation,performativeDebias,performativepower}, reinforcement learning \cite{performativeReinforcement,pollatos2025corruption}, and federated learning \cite{performativefederated}. A typical example is company bankruptcy prediction~\cite{performative}: Authorities develop a classification model based on company features to generate bankruptcy predictions. In response, companies modify their behavior, reflected in their updated features, to either avoid bankruptcy or further improve their financial status. This strategic adaptation may lead some companies to successfully avoid bankruptcy after altering their features, necessitating an update to their ground-truth label (from bankruptcy to solvency). Such performativity results in evolving data distributions, which can be formalized as a data distribution map $\mathcal{D}(\theta)$, with $\theta$ representing the model and $\mathcal{D}(\theta)$ the induced distribution.  

In performative prediction, the so-called \textit{performative loss (risk)} is defined as the expected loss $\ell(\mathbf{X},\mathbf{y},\theta)$ over the model-induced data distribution:
{
\begin{equation}
    \text{Performative Loss (Risk)} =  \underset{\{\mathbf{X},\mathbf{y}\}\sim\mathcal{D}(\theta)}{\operatorname*{\ell(\mathbf{X},\mathbf{y},\theta)}},
    \label{performativeloss}
\end{equation}}

\noindent where $\{\mathbf{X},\mathbf{y}\}$ are samples drawn from the distribution induced by model $\theta$. Given this loss function, performative prediction research primarily focuses on two objectives: performative stability and performative optimality~\cite{performative}.

A performatively stable model minimizes the performative loss in the data distribution it induces, thereby preventing further distribution shifts and avoiding the necessity for subsequent model updates. Following the pioneering work in~\cite{performative}, research on performative stability has concentrated on deriving and analyzing various conditions for convergence to this state~\cite{performative,stochasticperformative,performativestateful,beyondPerformative,multiplayer_performative}. In contrast, a performatively optimal model achieves the lowest possible performative loss across all potential induced distributions, although it may not be performatively stable. Studies in performative optimality either aim to construct convex performative loss functions for direct optimization~\cite{outside,zeroinequality,pluginperformative,performativebanditfeedback} or utilize gradient-based strategies to find the optimal solution~\cite{howtolearn,learngradually,TwotimescaleDerivative,zhu2023online}.

While these works provide valuable foundations for performative prediction, their theoretical analyses and methodological designs often rely on assumptions that may not always hold in real‐world applications. For instance, convergence analyses for performative stability typically assume that the gradient of the loss function is bounded with respect to data $\{\mathbf{X,y}\}$. Similarly, research exploring gradient‐based strategies for performative optimality often assumes a bounded gradient of the performative loss with respect to data $\{\mathbf{X,y}\}$ or model $\theta$. However, as noted in \cite{performativebanditfeedback}, gradient calculations may be complex or infeasible in some cases, potentially limiting the applicability of these methods in various domains. For works constructing convex performative loss, the assumption of loss convexity with respect to the model $\theta$ or the data distribution parameter $\varphi$ is common. However, as the performative loss integrates the data distribution map $\mathcal{D}(\theta)$ into the formulation, convexity cannot be guaranteed, especially when the format of $\mathcal{D}(\theta)$ is complex. 


To maintain consistency between theoretical guarantees and empirical validations, the aforementioned works often rely on linear cases in their examples and evaluations, where the assumptions of bounded gradients and convex loss functions can be easily satisfied. However, this linearity, while facilitating the illustration of concepts, may not hold in real-world applications where data typically exhibit complex nonlinear characteristics \cite{MAO2025111280,LI2024110551,wang2023local}. 
For instance, in the company bankruptcy prediction example, the relationship between a company's features (e.g., advertising spending) and its bankruptcy status cannot be simplified to a linear relation \cite{penman2013financial}. 
To the best of our knowledge, the first and only work addressing nonlinear performative prediction is proposed by~\cite{performativedeeplearning}, which develops a neural network model for this purpose. By constructing a convex loss function with respect to model prediction $f_\theta(x)$ rather than model parameters $\theta$, it avoids the difficulties associated with complex model parameters, thus guaranteeing convergence to performative stability for its nonlinear model. However, this method still relies on potentially uncontrollable conditions within its framework. For instance, it assumes a bounded loss gradient with respect to $f_\theta(x)$, which may not hold in complex applications or scenarios.

To bridge the gap between idealized assumptions and real-world complexities, this paper presents a novel approach to performative prediction. Our method is designed to handle both linear and nonlinear datasets, avoiding potentially constraining assumptions yet preserving essential theoretical properties. The key contributions of this work can be summarized as follows:

\begin{itemize}
    \item We introduce a novel sensitivity definition for the data distribution map, quantifying distribution shifts by measuring the discrepancy between prediction errors on two model-induced data distributions. This measurable quantity establishes a robust foundation for convergence analysis.

    \item Based on this new sensitivity definition, we derive conditions for performative stability in the proposed method. Our theoretical analysis shows that, with an appropriately selected model parameter—which is fully controllable by the user—the method converges to a performatively stable model at a linear rate. 
    
    \item Guided by our theoretical analysis, we design an algorithm for the proposed method that ensures the convergence to a performatively stable model in realistic scenarios where only finite samples are available.
    
    \item We conduct comprehensive experiments using both synthetic and real-world datasets, with linear and nonlinear cases across various data distribution maps. Comparison with six state-of-the-art approaches in performative prediction demonstrates the effectiveness of our method in terms of both accuracy and stability.
\end{itemize}

The rest of the paper is organized as follows. In Section~\ref{method}, we introduce the details of our method.
Experimental results are presented and discussed in Section~\ref{experiment}.
We conclude our paper in Section~\ref{conclusion}.

\section{Nonlinear Performative Prediction via Maximum Margin Strategy} \label{method}
In this section, we detail the proposed method, \textbf{N}onlinear \textbf{P}erformative \textbf{P}rediction via \textbf{M}aximum \textbf{M}argin strategy (NP$^2$M$^2$). First, we state the problem of nonlinear performative prediction and formulate our objective function. Then, we introduce a novel definition to describe the sensitivity of the data distribution map $\mathcal{D}(\theta)$, which is crucial for quantifying the magnitude of change in the data distribution $\mathcal{D}$ due to the change in the model $\theta$. Based on this, we derive conditions for performative stability from both theoretical and practical perspectives. 
Finally, we develop an algorithm grounded in our theoretical results, which ensures the performative stability of the predictive model.


\subsection{Problem Statement}
The problem of nonlinear performative prediction can be stated as follows. Given a dataset $\{\mathbf{\tilde{X},y}\}$, which is sampled from a nonlinear data distribution $\mathcal{D}(\theta)$ induced by a deployed model $\theta$, i.e., $\{\mathbf{\tilde{X},y}\} \sim \mathcal{D}(\theta)$, we seek to find a performatively stable model $\theta_{PS}$ that minimizes the loss in its induced distribution $\mathcal{D}(\theta_{PS})$. By doing so, the resulting model is optimal for the current data distribution without causing further distribution shift. Mathematically, our objective function can be formulated as follows\footnote{Note that in this paper, we use the classification task as an example to demonstrate and explain our idea. In fact, the proposed framework and the theoretical results of our work can be flexibly extended to other learning tasks, such as ranking and regression.}:
{
\begin{equation}
    \begin{aligned}
        \theta_{PS} &= \arg \min_\theta \underset{\{\mathbf{X},\mathbf{y}\}\sim\mathcal{D}(\theta_{PS})}{L(\theta)} \\
        &= \arg \min_\theta \frac{1}{2}\|\theta\|^2 + \frac{C}{n} \sum_{i=1}^n\left(\underset{\{\mathbf{X},\mathbf{y}\}\sim\mathcal{D}(\theta_{PS})}{\max(0,1-y_i \theta^T \phi(x_i))}\right),
    \end{aligned}
    \label{svmprime}
\end{equation}}

\noindent where $\|\theta\|^2$ is the regularization term maximizing the class margin, $\sum_{i}\max(0,1-y_i \theta^T \phi(x_i))$ is the empirical risk term minimizing the prediction error via hinge loss, $\phi(\cdot)$ maps the original data to a higher-dimensional, linearly separable space, $n$ denotes the number of data samples, $C$ is a balancing parameter, and $x_i$ represents an augmented vector derived from the $i^{th}$ original data $\tilde{x}_i$, which will be explained in detail later in this paragraph. In Eq.~(\ref{svmprime}), we employ the maximum margin approach to formulate our objective function due to its desirable property of convexity, which ensures that the optimization problem remains tractable and globally solvable \cite{wang2024fast,FONSECANETO2025111339,LU2024110753}. With the kernel function $K(x,x^\prime) = \phi(x)\cdot \phi(x^\prime)$ \cite{svm,svmbook}, data can be implicitly mapped into a higher-dimensional (possibly infinite-dimensional) feature space where linear separability may be achieved, avoiding explicitly performing the transformation \cite{guo2021feature,DAI2025111399,LEE2025111297}. Thus, we are able to handle complex nonlinear datasets in the original feature space, enhancing the applicability of the proposed method in various applications.
To guarantee the strong convexity of the learning problem, which is a required property for deriving the condition for performative stability, we augment the original feature space $\mathbf{\tilde{X}}$ to a new space $\mathbf{X}$ by adding one dimension, i.e., $x_i = (\tilde{x}_i,\tau)$, where $\tau$ is a fixed scalar value. As a result, the model parameters become $\theta = (\tilde{\theta},\frac{b}{\tau})$, and the decision function in our formulation is expressed as: $f(\tilde{x}_i) = \theta^T \phi(x_i) = \tilde{\theta}^T \phi(\tilde{x}_i) +b$ \cite{introductiontosvm,suykens2014regularization}. 

A natural approach to finding the solution to the optimization problem formulated in Eq.~(\ref{svmprime}) is to repeatedly minimize the risk over the model-induced data until convergence:
{
\begin{equation}
    \theta_{t+1} = G(\theta_t) = \arg \min_\theta \mathop{L(\theta,\phi(\mathbf{X}),\mathbf{y})}\limits_{\{\mathbf{X},\mathbf{y}\}\sim \mathcal{D}(\theta_t)}.
    \label{Gtheta}
\end{equation}}

\noindent In the subsequent subsections, we will analyze the convergence properties of the repeated risk minimization (RRM) procedure in Eq.~(\ref{Gtheta}) and derive the conditions under which this iterative process converges to a performatively stable model.

\subsection{Definition of \texorpdfstring{$\varepsilon$}{epsilon}-sensitivity for \texorpdfstring{NP$^2$M$^2$}{}}
To analyze the convergence properties of the RRM procedure, it is essential to quantify the performative effect, i.e., the sensitivity of the data distribution map $\mathcal{D}(\theta)$ with respect to $\theta$. However, directly measuring this effect--how model influences the data distribution shift--is challenging due to its complexity and the multitude of factors involved. To address this, we propose defining the sensitivity of $\mathcal{D}(\theta)$ in terms of prediction error. This idea is inspired by the fundamental concept that changes in data distribution correspond to changes in model performance; if two data distributions are similar, the errors produced by a predictive model on these two distributions should be comparable. Thus, measuring the discrepancy between prediction errors resulting from data distribution shifts serves as an effective indicator of the performative effect in the targeted learning problem. Accordingly, we define the $\varepsilon$-sensitivity for NP$^2$M$^2$ as follows:
\begin{definition} \label{sensitive}
    ($\varepsilon$-sensitivity for NP$^2$M$^2$). We define the data distribution map $\mathcal{D}(\theta)$ as $\varepsilon$-sensitive if the following inequality holds:  
    \begin{equation}
        \left|\frac{1}{n}\mathop{\sum_{i}}\limits_{y_i \theta^{*\top} \phi(x_i)<1}\mathop{y_i \theta^{\sharp\top} \phi(x_i)}\limits_{\{\mathbf{X},\mathbf{y}\}\sim \mathcal{D}(\theta)} - \frac{1}{n^\prime}\mathop{\sum_{i}}\limits_{y_i^\prime \theta^{*\top} \phi(x_i^\prime)<1}\mathop{y_i^\prime \theta^{\sharp \top}  \phi(x_i^\prime)}\limits_{\{\mathbf{X}^\prime,\mathbf{y}^\prime\}\sim \mathcal{D}(\theta^\prime)} \right| \le \varepsilon \|\theta - \theta^\prime\|,
        \label{sensitiveeq}
    \end{equation}
        
    
    \noindent where $\theta$ and $\theta^\prime$ represent two deployed models, and $\theta^\sharp$ and 
    $\theta^*$ are defined as: $\theta^\sharp = \frac{G(\theta^\prime)-G(\theta)}{\|G(\theta^\prime)-G(\theta)\|}$ and $\theta^* = G(\theta^\prime)$.
    

    
\end{definition}

The left‐hand side of the inequality represents the discrepancy between prediction errors produced by the same model on two distributions. Note that this model used for measuring prediction errors is not necessarily identical to the deployed models $\theta$ and $\theta^\prime$. The right-hand side of the inequality denotes the difference between deployed models, scaled by the parameter $\varepsilon$. This definition provides a measurable bound for the sensitivity of $\mathcal{D}(\theta)$ using the model difference, thereby establishing a foundation for convergence analysis. Note that the $\varepsilon$-sensitivity is defined in terms of the calculable empirical error regardless of the specific distribution type, and can be applied to all data distributions.


\subsection{Performative Stability of NP$^2$M$^2$}

With the \texorpdfstring{$\varepsilon$}{epsilon}-sensitivity introduced in Definition~\ref{sensitive}, we can now derive the condition for performative stability for NP$^2$M$^2$ in the following theorem.



\begin{theorem} \label{stable}
    If $\mathcal{D}(\theta)$ is $\varepsilon$-sensitive as introduced in Definition~\ref{sensitive}, then for the RRM procedure given in Eq.~(\ref{Gtheta}), the following holds:
    \begin{enumerate}
        \item {
        $\left\|G(\theta) - G(\theta^\prime)\right\| \le \varepsilon C \left\|\theta - \theta^\prime\right\|$}, and 
        
        \item If $\varepsilon C < 1$, the RRM procedure converges to a performatively stable model $\theta_{PS}$ at a linear rate:
    \end{enumerate}
    \vspace{-0.10cm}
    {%
    \begin{equation}
        \|\theta_{t} - \theta_{PS}\| \le \delta \ \text{ for } \ t \ge \frac{\log(\frac{\delta}{\|\theta_0-\theta_{PS}\|})}{\log(\varepsilon C)}.
    \end{equation}}
\end{theorem}

\begin{pf}
    The objective function of NP$^2$M$^2$ formulated in Eq.~(\ref{svmprime}) is 1-strongly convex with respect to $\theta$. According to the definition of $G(\theta)$ in Eq.~(\ref{Gtheta}), we have the following:    
    \begin{equation}%
        \begin{aligned}
        \mathop{L(G(\theta))}\limits_{\{\mathbf{X},\mathbf{y}\}\sim \mathcal{D}(\theta)} - \mathop{L(G(\theta^\prime))}\limits_{\{\mathbf{X},\mathbf{y}\}\sim \mathcal{D}(\theta)} \ge & (G(\theta)-G(\theta^\prime))^T \nabla_{G(\theta^\prime)} \mathop{L(G(\theta^\prime))}\limits_{\{\mathbf{X},\mathbf{y}\}\sim \mathcal{D}(\theta)} + \frac{1}{2}\|G(\theta) - G(\theta^\prime)\|^2,\\
        \mathop{L(G(\theta^\prime))}\limits_{\{\mathbf{X},\mathbf{y}\}\sim \mathcal{D}(\theta)} - \mathop{L(G(\theta))}\limits_{\{\mathbf{X},\mathbf{y}\}\sim \mathcal{D}(\theta)} \ge & \frac{1}{2}\|G(\theta) - G(\theta^\prime)\|^2,
        \end{aligned}
    \label{stronglyconvex}
    \end{equation}
    where $\nabla_{G(\theta^\prime)} \mathop{L(G(\theta^\prime))}\limits_{\{\mathbf{X},\mathbf{y}\}\sim \mathcal{D}(\theta)}$ is the gradient of $\mathop{L(G(\theta^\prime))}\limits_{\{\mathbf{X},\mathbf{y}\}\sim \mathcal{D}(\theta)}$ with respect to $G(\theta^\prime)$: 
    {%
    \begin{equation}
        \nabla_{G(\theta^\prime)} \mathop{L(G(\theta^\prime))}\limits_{\{\mathbf{X},\mathbf{y}\}\sim \mathcal{D}(\theta)} = G(\theta^\prime) - \frac{C}{n} \mathop{\sum_{i}}\limits_{y_i G(\theta^\prime)^T \phi(x_i) < 1} y_i \phi(x_i).
        \label{svmgradient}
    \end{equation}}
    
    \noindent Combining the two inequalities in Eq.~(\ref{stronglyconvex}), we obtain:
    {%
    \begin{equation}
        \|G(\theta) - G(\theta^\prime)\|^2 \le (G(\theta^\prime)-G(\theta))^T \nabla_{G(\theta^\prime)}  \mathop{L(G(\theta^\prime))}\limits_{\{\mathbf{X},\mathbf{y}\}\sim \mathcal{D}(\theta)}.
        \label{ine1}
    \end{equation}}
    
    \noindent According to the $\varepsilon$-sensitivity of $\mathcal{D}(\theta)$ in Definition~\ref{sensitive}, we derive:
    \begin{equation}%
        \begin{aligned}
            (G(\theta^\prime)-G(\theta))^T \left(\nabla_{G(\theta^\prime)}  \mathop{L(G(\theta^\prime))}\limits_{\{\mathbf{X},\mathbf{y}\}\sim \mathcal{D}(\theta)} - \nabla_{G(\theta^\prime)}  \mathop{L(G(\theta^\prime))}\limits_{\{\mathbf{X}^\prime,\mathbf{y}^\prime\}\sim \mathcal{D}(\theta^\prime)}\right) \le \varepsilon C \|\theta - \theta^\prime\|  \|G(\theta)-G(\theta^\prime)\|.\\
        \end{aligned}
        \label{longineq}
    \end{equation}      
    Since $L(\theta)$ is convex and $G(\theta^\prime)$ is the optimal solution for $\mathop{L(G(\theta^\prime))}\limits_{\{\mathbf{X}^\prime,\mathbf{y}^\prime\}\sim \mathcal{D}(\theta^\prime)}$ \cite{bubeck2015convex}, we have:
    {%
    \begin{equation}
        (G(\theta^\prime)-G(\theta))^T \nabla_{G(\theta^\prime)} \mathop{L(G(\theta^\prime))}\limits_{\{\mathbf{X}^\prime,\mathbf{y}^\prime\}\sim \mathcal{D}(\theta^\prime)} \le 0.
        \label{convex_and_optimal}
    \end{equation}}

\noindent Combining Eqs.~(\ref{longineq}-\ref{convex_and_optimal}), we get the following:
    {%
    \begin{equation}
        (G(\theta^\prime)-G(\theta))^T \nabla_{G(\theta^\prime)} \mathop{L(G(\theta^\prime))}\limits_{\{\mathbf{X},\mathbf{y}\}\sim \mathcal{D}(\theta)} \le \varepsilon C \|\theta - \theta^\prime\|  \|G(\theta)-G(\theta^\prime)\|.
        \label{ine3}
    \end{equation}}
    
    \noindent With both Eq.~(\ref{ine1}) and Eq.~(\ref{ine3}), we have:
    {%
    \begin{equation}
        \|G(\theta) - G(\theta^\prime)\|^2 \le \varepsilon C \|\theta - \theta^\prime\|  \|G(\theta)-G(\theta^\prime)\|.
    \end{equation}   } 
    
    \noindent Dividing both sides by $\|G(\theta) - G(\theta^\prime)\|$, we obtain the first inequality in Theorem~\ref{stable}:
    {%
    \begin{equation}
        \|G(\theta) - G(\theta^\prime)\| \le \varepsilon C \|\theta - \theta^\prime\|.
        \label{theorem1-point1}
    \end{equation}}

    If we let $\theta = \theta_{t-1}$ and $\theta^\prime = \theta_{PS}$, then we know that $G(\theta) = \theta_{t}$. Moreover, since $\theta^\prime$ is the performatively stable model $\theta_{PS}$, we have $G(\theta^\prime) = G(\theta_{PS})= \theta_{PS} = \theta^\prime$. Applying these to Eq.~(\ref{theorem1-point1}), we derive:
    {%
    \begin{equation}
        \|\theta_{t} - \theta_{PS}\| \le \varepsilon C \|\theta_{t-1} - \theta_{PS}\| \le ( \varepsilon C)^t \|\theta_{0} - \theta_{PS}\|.
    \end{equation}}
    
    \noindent When $\varepsilon C < 1$, we can further conclude: 
    {%
    \begin{equation}
        \|\theta_{t} - \theta_{PS}\| \le \delta \  for \ t \ge \frac{\log(\frac{\delta}{\|\theta_0-\theta_{PS}\|})}{\log(\varepsilon C)}.
    \end{equation}}
    $\qed$
\end{pf}


\subsection{Finite-Sample Analysis for NP$^2$M$^2$}
In Theorem~\ref{stable}, $\varepsilon$ is measurable, and $C$ is the user-controllable parameter that can be flexibly adjusted, which is crucial for ensuring the performative stability of predictive models in practical applications.
However, this theorem operates under the assumption of full accessibility to the data distribution, which may not always be satisfied in real-world scenarios.
Thus, we extend our analysis to derive the conditions for performative stability of the proposed method in the more realistic context where only finite samples can be collected. 

\begin{theorem}\label{theorem2}
    If $\mathcal{D}(\theta)$ is $\varepsilon$-sensitive as introduced in Definition~\ref{sensitive} and $2\varepsilon C < 1$, then for $t \geq \log(\frac\delta{\|\theta_0-\theta_{PS}\|}) /\log(\varepsilon C)$, the RRM procedure defined in Eq.~(\ref{Gtheta}), when applied to the empirical dataset, satisfies:
    \begin{equation}
        \|\theta_t - \theta_{PS}\|\le  2 \varepsilon C \|\theta_{t-1} - \theta_{PS}\| \le (2 \varepsilon C)^t  \|\theta_{0} - \theta_{PS}\| \le \delta,
        \label{eqtheorem2}
    \end{equation}
    
    \noindent with the probability $2\Phi(\frac{\varepsilon \delta\sqrt{n_t}}{\sigma})-1$, where $\Phi(\cdot)$ is the standard normal cumulative distribution function, $\delta$ is a parameter controlling the convergence rate, $n_t$ denotes the size of the empirical dataset at time step $t$, and $\sigma$ represents the standard deviation of the prediction error measured by the model $\theta^\sharp$ as defined in Definition~\ref{sensitive}.
\end{theorem}

\begin{pf}
    We denote $\overline{\xi} = \mathop{\mathbb{E}}\limits_{y_i \theta^{*\top}\phi(x_i)<1}  y_i\theta^{\sharp  \top}\phi(x_i)$ as the expectation value of the term in Eq.~(\ref{sensitiveeq}), $\xi_i = y_i \theta^{\sharp \top} \phi(x_i)$ as the error value for the $i^{th}$ data $\{x_i,y_i\}$ that satisfies $y_i \theta^{*\top}\phi(x_i)<1$, and $\overline{\xi}_n= \frac{1}{n}\mathop{\sum_i}\limits_{y_i \theta^{*\top}\phi(x_i)<1} y_i \theta^{\sharp \top}  \phi(x_i)$ as the average error value with $n$ samples. Here, $\theta^\sharp$ is a unit model with $\|\theta^\sharp\| = 1$. 
    The variance of $\xi_i$ can be calculated as $Var\left[\xi_i\right] = \sigma^2 = E(\xi_i - \overline{\xi}_n)^2$.
	
    We now calculate the discrepancy between the average $\overline{\xi}_n$ and the real expectation $\overline{\xi}$ according to the Central Limit Theorem \cite{bauer2001measure}. For any real number $z$, we have:
    \begin{equation}%
        P\left[|\overline{\xi}_n - \overline{\xi}| \le \frac{z}{\sqrt{n}}\right] = 2\Phi(\frac{z}{\sigma})-1.
    \end{equation}
    
    Let $\delta = \frac{z}{\sqrt{n}}$, then $\frac{z}{\sigma} = \frac{\delta\sqrt{n}}{\sigma}$.	By sampling $n$ data, we have $|\overline{\xi}_n - \overline{\xi}| \le \delta$ with probability $2\Phi(\frac{\delta\sqrt{n}}{\sigma})-1$.
    
    We then prove the theorem by investigating two cases: 
    \begin{itemize}
        \item Case 1: When $\|\theta_t - \theta_{PS}\| > \delta$, $2\varepsilon C < 1$, we prove that, by performing the RRM procedure on the empirical dataset, we have $\|\theta_t - \theta_{PS}\| \le 2\varepsilon C \|\theta_{t-1} - \theta_{PS}\|$ with probability $2\Phi(\frac{\varepsilon \delta\sqrt{n_t}}{\sigma})-1$.
        
        \item Case 2: When $\|\theta_t - \theta_{PS}\| \le \delta$, $2\varepsilon C < 1$, we prove that, by performing the RRM procedure on the empirical dataset, we have $\|\theta_t - \theta_{PS}\| \le \delta$ with probability $2\Phi(\frac{\varepsilon \delta\sqrt{n_t}}{\sigma})-1$.
    \end{itemize}
    
    \underline{Case 1:} If the current model $\theta_{t}$ is far from the performatively stable model $\theta_{PS}$, we show that with a probability the model in the next iteration $G_{n_t}(\theta_{t})$ will move towards $\theta_{PS}$. 
    
    We can apply the triangle inequality:
    \begin{equation}%
        |\overline{\xi_{n_t}} - \overline{\xi_{PS}}| \le |\overline{\xi_{n_t}} - \overline{\xi_{t}}| + |\overline{\xi_{t}} - \overline{\xi_{PS}}|,
        \label{triangle}
    \end{equation}
    where $\overline{\xi_{n_t}}$ is the average over $n_t$ samples in time step $t$, $\overline{\xi_{t}}$ is the real expectation of the underlying distribution in time step $t$, and $\overline{\xi_{PS}}$ is the real expectation of the performatively stable data distribution $\mathcal{D}(\theta_{PS})$.
    
    From Definition \ref{sensitive}, we have $ |\overline{\xi_{t}} - \overline{\xi_{PS}}| \le \varepsilon \|\theta_t - \theta_{PS}\|$. According to the Central Limit Theorem \cite{bauer2001measure}, we have $|\overline{\xi_{n_t}} - \overline{\xi_{t}}|\le \varepsilon \delta$ with probability $2\Phi(\frac{\varepsilon \delta \sqrt{n_t}}{\sigma})-1$.
    
    Then, we have
    \begin{equation}%
        |\overline{\xi_{n_t}} - \overline{\xi_{PS}}| \le |\overline{\xi_{n_t}} - \overline{\xi_{t}}| + |\overline{\xi_{t}} - \overline{\xi_{PS}}| 
        \le \varepsilon \delta + \varepsilon \|\theta_t - \theta_{PS}\|\le 2 \varepsilon \|\theta_t - \theta_{PS}\|,
        \label{ine4}
    \end{equation}
    with probability $2\Phi(\frac{\varepsilon \delta\sqrt{n_t}}{\sigma})-1$.
    
    Due to the convexity of the loss function in Eq.~(\ref{svmprime}), we have: 
    \begin{equation}%
            \mathop{L_{n_t}(G(\theta_{PS}))}\limits_{\{\mathbf{X},\mathbf{y}\}\sim \mathcal{D}_{n_t}(\theta_t)} -\mathop{L_{n_t}(G_{n_t}(\theta_t))}\limits_{\{\mathbf{X},\mathbf{y}\}\sim \mathcal{D}_{n_t}(\theta_t)} \ge (G(\theta_{PS}) - G_{n_t}(\theta_t))^T \nabla_{G_{n_t}(\theta_t)} \mathop{L(G_{n_t}(\theta_t))}\limits_{\{\mathbf{X},\mathbf{y}\}\sim \mathcal{D}_{n_t}(\theta_t)} \ge 0,
    \end{equation}
    and
    \begin{equation}%
            \mathop{L(G_{n_t}(\theta_t))}\limits_{\{\mathbf{X},\mathbf{y}\}\sim \mathcal{D}(\theta_{PS})} - \mathop{L(G(\theta_{PS}))}\limits_{\{\mathbf{X},\mathbf{y}\}\sim \mathcal{D}(\theta_{PS})} 
            \ge ( G_{n_t}(\theta_t)-G(\theta_{PS}))^T \nabla_{G(\theta_{PS})}\mathop{L(G(\theta_{PS}) )}\limits_{\{\mathbf{X},\mathbf{y}\}\sim \mathcal{D}(\theta_{PS})} \ge 0,
    \end{equation}
    where $n_t$ is the number of data that sampled in time step $t$, $L_{n_t}(\cdot)$ is the loss value with $n_t$ samples, $\mathcal{D}_{n_t}(\theta_t)$ is the set of the $n_t$ data that we sampled from the data distribution $\mathcal{D}(\theta_t)$, and $G_{n_t}(\theta_t)$ is the optimal solution of the loss function $\mathop{L_{n_t}(\cdot)}\limits_{\{\mathbf{X},\mathbf{y}\}\sim \mathcal{D}_{n_t}(\theta_t)}$.
    
    Combining the above two inequalities, we have 
    \begin{equation}%
    \begin{aligned}
    &(G(\theta_{PS}) - G_{n_t}(\theta_t))^T \nabla_{G_{n_t}(\theta_t)}\mathop{L(G_{n_t}(\theta_t))}\limits_{\{\mathbf{X},\mathbf{y}\}\sim \mathcal{D}_{n_t}(\theta_t)} \\
    + &( G_{n_t}(\theta_t)-G(\theta_{PS}))^T \nabla_{G(\theta_{PS})} \mathop{L(G(\theta_{PS}) )}\limits_{\{\mathbf{X},\mathbf{y}\}\sim \mathcal{D}(\theta_{PS})} \ge 0.
    \end{aligned}
    \end{equation}   
    Then, we have:
    \begin{equation}%
        \begin{aligned}
            &(G_{n_t}(\theta_t) - G(\theta_{PS}))^T \left( \nabla_{G(\theta_{PS})} \mathop{L(G(\theta_{PS}) )}\limits_{\{\mathbf{X},\mathbf{y}\}\sim \mathcal{D}(\theta_{PS})}-\nabla_{G_{n_t}(\theta_t)} \mathop{L(G_{n_t}(\theta_t))}\limits_{\{\mathbf{X},\mathbf{y}\}\sim \mathcal{D}_{n_t}(\theta_t)} \right) \\
            = & (G_{n_t}(\theta_t) - G(\theta_{PS}))^T \left(\nabla_{G(\theta_{PS})}\mathop{L(G(\theta_{PS}) )}\limits_{\{\mathbf{X},\mathbf{y}\}\sim \mathcal{D}(\theta_{PS})} - \nabla_{G_{n_t}(\theta_t)} \mathop{L(G_{n_t}(\theta_t))}\limits_{\{\mathbf{X},\mathbf{y}\}\sim \mathcal{D}(\theta_{PS})} \right) \\
            &+ ( G_{n_t}(\theta_t)-G(\theta_{PS}))^T \left(\nabla_{G_{n_t}(\theta_t)}\mathop{L(G_{n_t}(\theta_t))}\limits_{\{\mathbf{X},\mathbf{y}\}\sim \mathcal{D}(\theta_{PS})}-\nabla_{G_{n_t}(\theta_t)} \mathop{L(G_{n_t}(\theta_t))}\limits_{\{\mathbf{X},\mathbf{y}\}\sim \mathcal{D}_{n_t}(\theta_t)} \right)\ge 0.
        \end{aligned}
    \end{equation}
    Therefore, we have
    \begin{equation}
    \begin{aligned}
        &(G_{n_t}(\theta_t) - G(\theta_{PS}))^T \left(\nabla_{G_{n_t}(\theta_t)}\mathop{L(G_{n_t}(\theta_t))}\limits_{\{\mathbf{X},\mathbf{y}\}\sim \mathcal{D}(\theta_{PS})} -\nabla_{G_{n_t}(\theta_t)} \mathop{L(G_{n_t}(\theta_t))}\limits_{\{\mathbf{X},\mathbf{y}\}\sim \mathcal{D}_{n_t}(\theta_t)}\right) \\
        \ge &( G_{n_t}(\theta_t)-G(\theta_{PS}))^T \left( \nabla_{G_{n_t}(\theta_t)} \mathop{L(G_{n_t}(\theta_t))}\limits_{\{\mathbf{X},\mathbf{y}\}\sim \mathcal{D}(\theta_{PS})}-\nabla_{G(\theta_{PS})}\mathop{L(G(\theta_{PS}) )}\limits_{\{\mathbf{X},\mathbf{y}\}\sim \mathcal{D}(\theta_{PS})}\right).
    \end{aligned}\label{eq28}
    \end{equation}
    Let $\theta^\sharp = \frac{G_{n_t}(\theta_t)-G(\theta_{PS})}{\|G_{n_t}(\theta_t)-G(\theta_{PS})\|}$, according to Eq.~(\ref{ine4}), with probability $2\Phi(\frac{\varepsilon \delta \sqrt{n_t}}{\sigma})-1$, we have : 
    \begin{equation}
    \begin{aligned}
        &(G_{n_t}(\theta_t) - G(\theta_{PS}))^T \left(\nabla_{G_{n_t}(\theta_t)} \mathop{L(G_{n_t}(\theta_t))}\limits_{\{\mathbf{X},\mathbf{y}\}\sim \mathcal{D}(\theta_{PS})} -\nabla_{G_{n_t}(\theta_t)} \mathop{L(G_{n_t}(\theta_t))}\limits_{\{\mathbf{X},\mathbf{y}\}\sim \mathcal{D}_{n_t}(\theta_t)}\right)\\
        \le &\left|(G_{n_t}(\theta_t) - G(\theta_{PS}))^T \left(\nabla_{G_{n_t}(\theta_t)}\mathop{L(G_{n_t}(\theta_t))}\limits_{\{\mathbf{X},\mathbf{y}\}\sim \mathcal{D}_{n_t}(\theta_t)} - \nabla_{G_{n_t}(\theta_t)} \mathop{L(G_{n_t}(\theta_t))}\limits_{\{\mathbf{X},\mathbf{y}\}\sim \mathcal{D}(\theta_{PS})}\right)\right| \\
        = &C|\overline{\xi_{n_t}} - \overline{\xi_{PS}}|  \|G_{n_t}(\theta_t) - G(\theta_{PS})\|\\
        \le &  2 \varepsilon C \|\theta_t - \theta_{PS}\| \|G_{n_t}(\theta_t) - G(\theta_{PS})\|.
    \end{aligned}
    \label{ine5}
    \end{equation}    
    Similar to the proof in Theorem \ref{stable}, because the loss function $L(\theta)$ in Eq.~(\ref{svmprime}) is 1-strongly convex, we have:
    {
    \begin{equation}
        \begin{aligned}
            &\mathop{L(G_{n_t}(\theta_t))}\limits_{\{\mathbf{X},\mathbf{y}\}\sim \mathcal{D}(\theta_{PS})} - \mathop{L(G(\theta_{PS}) )}\limits_{\{\mathbf{X},\mathbf{y}\}\sim \mathcal{D}(\theta_{PS})} \\
            \ge &(G_{n_t}(\theta_t) - G(\theta_{PS}))^T \mathop{\nabla_{G(\theta_{PS})} L(G(\theta_{PS}))}\limits_{\{\mathbf{X},\mathbf{y}\}\sim \mathcal{D}(\theta_{PS})} +\frac{1}{2}\|G_{n_t}(\theta_t) - G(\theta_{PS})\|^2 ,
            \end{aligned}
    \end{equation}}
    and
    {
    \begin{equation}
    \hspace{-0.5cm}
        \begin{aligned}
        &\mathop{L(G(\theta_{PS}) )}\limits_{\{\mathbf{X},\mathbf{y}\}\sim \mathcal{D}(\theta_{PS})} - \mathop{L(G_{n_t}(\theta_t))}\limits_{\{\mathbf{X},\mathbf{y}\}\sim \mathcal{D}(\theta_{PS})} \\
        \ge & (G(\theta_{PS}) - G_{n_t}(\theta_t))^T \mathop{\nabla_{G_{n_t}(\theta_t)} L(G_{n_t}(\theta_t))}\limits_{\{\mathbf{X},\mathbf{y}\}\sim \mathcal{D}(\theta_{PS})} + \frac{1}{2}\|G_{n_t}(\theta_t) - G(\theta_{PS})\|^2.
        \end{aligned}
    \end{equation}}    
    Combining the above two inequalities, we have
    \begin{equation}
    \begin{aligned}
        &(G_{n_t}(\theta_t)-G(\theta_{PS}))^T \left(\nabla_{G_{n_t}(\theta_t)}\mathop{L(G_{n_t}(\theta_t))}\limits_{\{\mathbf{X},\mathbf{y}\}\sim \mathcal{D}(\theta_{PS})}-\nabla_{G(\theta_{PS})}\mathop{L(G(\theta_{PS}) )}\limits_{\{\mathbf{X},\mathbf{y}\}\sim \mathcal{D}(\theta_{PS})}\right)\\
        \ge &\|G_{n_t}(\theta_t) - G(\theta_{PS})\|^2.
    \end{aligned}
        \label{ine6}
    \end{equation}    
    Combining Eqs.~(\ref{eq28}), (\ref{ine5}),~and~(\ref{ine6}), we have:
    {
    \begin{equation}
        \begin{aligned}
            &\|G_{n_t}(\theta_t) - G(\theta_{PS})\|^2  \le 2 \varepsilon C \|\theta_t - \theta_{PS}\| \|G_{n_t}(\theta_t) - G(\theta_{PS})\|,
        \end{aligned}
    \end{equation}}
    
    \noindent with probability $2\Phi(\frac{\varepsilon \delta\sqrt{n_t}}{\sigma})-1$.
    Dividing both sides by $\|G_{n_t}(\theta_t) - G(\theta_{PS})\| $, we obtain:
    \begin{equation}
    \|G_{n_t}(\theta_t) - G(\theta_{PS})\| \le 2 \varepsilon C \|\theta_t - \theta_{PS}\|,
    \end{equation}
    with probability $2\Phi(\frac{\varepsilon \delta\sqrt{n_t}}{\sigma})-1$.
    
    As a result, we conclude that the RRM procedure on empirical dataset will move towards the performatively stable model $\theta_{PS}$ at each iteration with probability $2\Phi(\frac{\varepsilon \delta\sqrt{n_t}}{\sigma})-1$ when $\|\theta_t - \theta_{PS}\| > \delta$. Therefore, for $t\geq\frac{\log\left(\frac\delta{\|\theta_0-\theta_{PS}\|}\right)}{\log(\varepsilon C)}$, we have:   \begin{equation}\hspace{-0.2cm}
        \|\theta_t - \theta_{PS}\|\le  2 \varepsilon C \|\theta_{t-1} - \theta_{PS}\| \le (2 \varepsilon C)^t  \|\theta_{0} - \theta_{PS}\| \le \delta,
    \end{equation}
    with probability $2\Phi(\frac{\varepsilon \delta\sqrt{n_t}}{\sigma})-1$.

    \vspace{0.5cm}
    
    \underline{Case 2: } We show that the RRM procedure on empirical dataset stays $\|\theta_t - \theta_{PS}\|\le \delta$ with a probability. 

    From Definition \ref{sensitive}, we have $ |\overline{\xi_{t}} - \overline{\xi_{PS}}| \le \varepsilon \|\theta_t - \theta_{PS}\| \le \varepsilon \delta$. Similar to Case 1, we apply the triangle inequality :
    \begin{equation}
        |\overline{\xi_{n_t}} - \overline{\xi_{PS}}| \le |\overline{\xi_{n_t}} - \overline{\xi_{t}}| + |\overline{\xi_{t}} - \overline{\xi_{PS}}| \le |\overline{\xi_{n_t}} - \overline{\xi_{t}}| + \varepsilon \delta.
    \end{equation}

    According to the Central Limit Theorem \cite{bauer2001measure}, we have $|\overline{\xi_{n_t}} - \overline{\xi_{t}}|\le \varepsilon \delta$ with probability $2\Phi(\frac{\varepsilon \delta \sqrt{n_t}}{\sigma})-1$. As a result, we know that the following inequality holds: 
    \begin{equation}
        |\overline{\xi_{n_t}} - \overline{\xi_{PS}}| \le 2 \varepsilon \delta,
        \label{eq37}
    \end{equation}
    with probability $2\Phi(\frac{\varepsilon \delta\sqrt{n_t}}{\sigma})-1$. By replacing the Eq.~(\ref{ine4}) in Case 1 with the above Eq.~(\ref{eq37}) and applying the same procedure used in the proof of Case 1, 
    we can conclude that
    \begin{equation}
        \|\theta_{t} - \theta_{PS}\|\le 2 \varepsilon C \delta \le \delta,
    \end{equation}
    with probability $2\Phi(\frac{\varepsilon \delta\sqrt{n_t}}{\sigma})-1$.
    $\qed$
\end{pf}

This theorem demonstrates that if the condition $2\varepsilon C < 1$ is met (which can be easily satisfied by appropriately selecting $C$ based on the measured $\varepsilon$), and given a sufficiently large sample size $n_t$ in each iteration, the RRM procedure is guaranteed to converge to a model approximating the performative stable model $\theta_{PS}$ with high probability.  


\subsection{Algorithm for NP$^2$M$^2$}
In this subsection, we present an iterative algorithm for the proposed NP$^2$M$^2$, ensuring the performative stability of the predictive model based on Theorem~\ref{theorem2}.
In each iteration, we first set a proper value of $C$ to satisfy $C < \frac{1}{2\varepsilon}$. With this balancing parameter, we obtain the optimal model for the current iteration by solving the optimization problem in Eq.~(\ref{Gtheta}). After that, we collect the data from the model-induced distribution to calculate $\varepsilon$, as in lines 5 and 12 of the above algorithm. This process repeats until convergence. Algorithm~\ref{method1} summarizes the entire procedure.

\begin{algorithm}[!t]
	\SetKwInOut{Input}{Input}
        \SetKwInOut{Output}{Output}
	
	\Input{Initial dataset $\{\mathbf{X}_0,\mathbf{y}_0\}\sim \mathcal{D}_0$, and $\alpha \in (0,0.5)$} 

        \Output{Performatively stable model $\theta_{PS}$}
	
	\BlankLine 

        $t \leftarrow 1$;
	
	Initialize $C$;
        
        $\theta_1 \leftarrow \arg \min_\theta \mathop{L(\theta)}\limits_{\{\mathbf{X}_{0},\mathbf{y}_{0}\}\sim \mathcal{D}_0}$;

        Collect data $\{\mathbf{X}_{1},\mathbf{y}_{1}\}\sim D(\theta_1)$;

        $\varepsilon_1 \leftarrow \frac{\left|\frac{1}{n_0}\sum\limits_{i=1}^{n_0} \mathop{y_i \theta_1^T\phi(x_i)}\limits_{\{\mathbf{X}_0,\mathbf{y}_0\}\sim \mathcal{D}_0} - \frac{1}{n_1}\sum\limits_{i=1}^{n_1}\mathop{y_i \theta_1^T\phi(x_i)}\limits_{\{\mathbf{X}_1,\mathbf{y}_1\}\sim \mathcal{D}(\theta_1)}\right|}{\|\theta_1\|^2 }$;

        $\bar{\varepsilon} \leftarrow \varepsilon_1$;

	\While{not converged}{
         $t \leftarrow t + 1$;
    
         $C \leftarrow \alpha / \bar{\varepsilon}$;\\
            $\theta_t \leftarrow \arg \min_\theta \mathop{L(\theta)}\limits_{\{\mathbf{X}_{t-1},\mathbf{y}_{t-1}\}\sim \mathcal{D}(\theta_{t-1})}$;\\
		
		Collect data $\{\mathbf{X}_{t},\mathbf{y}_{t}\}\sim D(\theta_t)$;
		
            \resizebox{0.9\linewidth}{!}{%
            $\varepsilon_t \leftarrow \frac{\left|\frac{1}{n_{t-1}}\sum\limits_{y_i \theta_{t-1}^T\phi(x_i)<1} \mathop{y_i (\theta_{t-1} - \theta_t)^T\phi(x_i)}\limits_{\{\mathbf{X}_{t-1},\mathbf{y}_{t-1}\}\sim \mathcal{D}(\theta_{t-1})} - \frac{1}{n_t}\sum\limits_{y_i \theta_{t-1}^T \phi(x_i)<1}\mathop{y_i (\theta_{t-1} - \theta_t)^T\phi(x_i)}\limits_{\{\mathbf{X}_t,\mathbf{y}_t\}\sim \mathcal{D}(\theta_t)}\right|}{\|\theta_{t-1} - \theta_t\|^2 }$};
		
		$\bar{\varepsilon} \leftarrow \frac{1}{t}\sum_{i=1}^t\varepsilon_i$;
	} 

        $\theta_{PS} \leftarrow \theta_t$;
	\caption{Nonlinear Performative Prediction via Maximum Margin approach (NP$^2$M$^2$)}
	\label{method1}
\end{algorithm}

\section{Empirical Evaluations}\label{experiment}
In this section, we evaluate the performance of the proposed method by comparing it with six state-of-the-art approaches on both synthetic and real-world datasets\footnote{The source code of our method is available at: `github.com/zhong-gz/NP2M2'.}.
First, we introduce our experimental setup, including baseline methods, parameter settings, evaluation metrics, and experimental procedure. Subsequently, we present and analyze the performance results of all methods on synthetic and real-world datasets.


\subsection{Experimental Settings}
\subsubsection{Baselines}

The six state-of-the-art approaches selected for comparison are: RRM with Logistic Regression (RRM LR) \cite{performative}, RGD with Logistic Regression (RGD LR) \cite{performative}, RRM with Neural Network (RRM NN) \cite{performativedeeplearning}, Two-Stage Approach (TSA) \cite{outside}, PerfGD \cite{howtolearn}, and DFO($\lambda$) \cite{TwotimescaleDerivative}. 
The details of these six baselines are given as follows\footnote{The baselines implementations can be found in 
`github.com/jcperdomo/performative-prediction' \cite{performative},
`github.com/zleizzo/PerfGD' \cite{howtolearn},
`proceedings.mlr.press/v139/miller21a/miller21a-supp.zip' \cite{outside},
`github.com/mhrnz/Performative-Prediction-with-Neural-Networks' \cite{performativedeeplearning},
and `github.com/Qiang-CU/DFO' \cite{TwotimescaleDerivative}.}.

The first three baselines are proposed for performative stability. 
	
\begin{itemize}	
    \item Repeat Risk Minimization with Logistic Regression (RRM LR) \cite{performative} trains and deploys logistic regression model repeatedly on current data distribution. This method ensures the performative stability, when $\varepsilon < \frac{\gamma}{\beta}$, where $\gamma$ is for the $\gamma$-strong convexity of the loss function, $\beta$ is for the $\beta$-joint smoothness of the loss function, and $\varepsilon$ is for the $\varepsilon$-sensitivity with respect to $\mathcal{W}_1$ distance of data distribution.

    \item Repeat Gradient Descent with Logistic Regression (RGD LR) \cite{performative} applies gradient descent to optimize the logistic regression on current data distribution. RGD LR ensures the performative stability, when $\varepsilon < \frac{\gamma}{(\beta+\gamma)(1+1.5\eta \beta)}$, where $\eta$ is the learning rate of gradient descent.

    \item RRM with Neural Networks (RRM NN) \cite{performativedeeplearning} trains and deploys neural networks on current data distribution. Different with RRM LR \cite{performative}, this method ensures the performative stability when $\mathcal{D}(\theta)$ is $\varepsilon$-sensitive with respect to Pearson $\mathcal{X}^2$ divergence, and the loss function $\ell(\theta)$ is strongly convex with respect to predictions and has bounded gradient norm.
\end{itemize}

The latter three baselines are proposed for performative optimality. 

\begin{itemize}
    \item Two-Stage Approach (TSA) \cite{outside} learns the data distribution map during the iterations. Then, the learned data distribution map $\mathcal{D}(\theta)$ is plugged into the logistic regression loss function $\ell(\mathcal{D}(\theta),\theta)$ for optimization. This method ensures the performative optimality when data distribution map $\mathcal{D}(\theta)$ satisfies mixture dominance with known form.
    
    \item PerfGD \cite{howtolearn} calculates the gradient of performative loss $\ell(\theta)$ by combining the gradient of logistic regression loss function $\ell(\theta)$ and the gradient induced by data distribution map $\mathcal{D}(\theta)$. PerfGD converges to performative optimality when its assumptions (e.g., the convexity of performative loss) are satisfied.

    \item DFO($\lambda$) \cite{TwotimescaleDerivative} estimates the gradient of performative loss with derivative-free optimization. The performative optimality can be achieved when the performative loss is convex. For non-convex performative loss, the resulting model leads to stationary points of performative loss when the loss value is bounded, and $\mathcal{D}(\theta)$ is $L_1$-Lipschitz.

\end{itemize}

\subsubsection{Parameter Setting}
For consistency with prior work, we set 
the learning rate $\lambda = 0.1$ for RGD LR, TSA, PerfGD, and RRM NN~\cite{performative,outside,performativedeeplearning}. The window size in PerfGD is set to $H = 2$~\cite{howtolearn}. The forgotten factor in DFO($\lambda$) is set to $\lambda = 0.5$~\cite{TwotimescaleDerivative}. The architecture of RRM NN consists of two hidden layers with eight neurons each, using the square loss function~\cite{performativedeeplearning}.
For the proposed NP$^2$M$^2$, we set $\alpha = 0.49$ to ensure that $\varepsilon C < \frac{1}{2}$ (Theorem \ref{theorem2}). In fact, our method demonstrates robustness to the variation of the parameter $\alpha$, which will be discussed in detail in Section \ref{ParameterSensitivity}.

\subsubsection{Evaluation Metrics}
We evaluate the performance of each method using the accuracy on given datasets. To assess the stability of methods, we introduce the model consistency, defined as the cosine similarity between models in consecutive iterations.
In the following, the details on both criteria, the accuracy and the model consistency, are given.

\begin{itemize}
    \item Accuracy:
    \begin{equation}
        \text{Accuracy} = \frac{\text{Number of correct prediction}}{\text{Total number of prediction}}.
    \end{equation}

    \item Previous research~\cite{performative,stochasticperformative} has used the model gap $\|\theta_t - \theta_{t-1}\|$ to evaluate the stability of methods, with a small gap indicating a stable model. While this criterion is suitable for assessing the stability in some scenarios, it is scale-sensitive. For example, in a classification decision function $f(x) = \text{sign}(\theta x)$, models $\theta$ and $2\theta$ produce identical classification results for all data and should be considered equivalent. However, their model gap $|\theta - 2\theta| = |\theta|$ is non-zero, potentially misrepresenting their relative stability. To avoid the influence caused by model scales, we apply cosine similarity as the metric to evaluate the model consistency. As a result, the range of model consistency is $\left[-1,1\right]$, where $-1$ indicates two completely opposite models, and $1$ indicates two identical models. The definition of model consistency between two consecutive iterations is given as follows:
    \begin{equation}
        \text{Model Consistency} = \frac{\theta_t \cdot \theta_{t-1}}{\|\theta_t\|\|\theta_{t-1}\|}.
    \end{equation}
    
    For different methods, the constructions of $\theta$ and calculations of model consistency are different, which is shown as follows.
    
    \begin{itemize}
        \item For NP$^2$M$^2$, $\theta = \sum_{i=1}^n \alpha_i y_i \phi(x_i)$ is the solution to the optimization problem in Eq.~(\ref{svmprime}). In nonlinear cases, the model consistency is calculated using the kernel function as follows:
        \begin{equation}
        \begin{aligned}
            &\text{Model Consistency} = \frac{\theta_t \cdot \theta_{t-1}}{\|\theta_t\|\|\theta_{t-1}\|} \\
            =& \frac{\mathbf{\alpha}_{t} \mathbf{y}_t K(\mathbf{X}_t,\mathbf{X}_{t-1})\mathbf{y}_{t-1}\mathbf{\alpha}_{t-1}} {\sqrt{\mathbf{\alpha}_{t} \mathbf{y}_t K(\mathbf{X}_t,\mathbf{X}_{t})\mathbf{y}_{t}\mathbf{\alpha}_{t}} \cdot \sqrt{\mathbf{\alpha}_{t-1} \mathbf{y}_{t-1} K(\mathbf{X}_{t-1},\mathbf{X}_{t-1})\mathbf{y}_{t-1}\mathbf{\alpha}_{t-1}}},
        \end{aligned}
        \end{equation}
        where $\alpha_t,\alpha_{t-1}$ are the Lagrange multipliers at time step $t,t-1$, and $K(\cdot,\cdot) = \phi(\cdot) \cdot \phi(\cdot)$ is the kernel function.
    
        \item For logistic regression based method (RRM  with Logistic Regression, RGD  with Logistic Regression, TSA, PerfGD), $\theta$ is constructed as:
        \begin{equation}
            \theta = \begin{bmatrix}
         w^T, b
        \end{bmatrix}^T,
        \end{equation}
        where $w$ is the weight vector of logistic regression and $b$ is the bias term of logistic regression.
    
        \item For RRM with Neural Networks, $\theta$ is constructed with the weight of all neurons: 
        \begin{equation}
            \theta = \begin{bmatrix}
            \theta_1^T, \theta_2^T, \cdots,     \theta_n^T
            \end{bmatrix}^T.
        \end{equation}
    \end{itemize}

\end{itemize}

\subsubsection{Experimental Procedure}
First, we collect initial data $\{\mathbf{X}_{0},\mathbf{y}_{0}\}$ from the base distribution $\mathcal{D}_0$. At each time step $t$, we train and deploy model $\theta_t$, then collect new data $\{\mathbf{X}_{t},\mathbf{y}_{t}\}$ from the distribution $\mathcal{D}(\theta_t)$. We evaluate the accuracy of model $\theta_t$ on $\{\mathbf{X}_{t},\mathbf{y}_{t}\}$ and calculate model consistency using $\theta_{t-1}$ and $\theta_{t}$. In our experiment, the procedure is performed for $100$ iterations ($T = 100$). For each setting, we repeat our experiments for $10$ times and report the average accuracy and model consistency for all methods after 20 iterations, allowing for initial convergence and stabilization of the models.

\subsection{Experiments with Synthetic Databases}
In this subsection, we evaluate the performance of the proposed method using both linear and nonlinear synthetic datasets, with two distinct data distribution maps $\mathcal{D}(\theta)$.

\subsubsection{Datasets}
For the linear dataset, we utilize the `make\_classification' function from Python's sklearn-library to generate a binary classification dataset. Each class contains $100$ samples with $50$ dimensions, of which $40$ dimensions are informative, and $10$ are noise dimensions.
For the nonlinear dataset, we use the `make\_circles' function from sklearn-library with a noise parameter of $0.2$ to generate a nonlinear binary classification dataset. Each class contains $250$ samples in $2$ dimensions.

\subsubsection{Data Distribution Maps}
We employ two data distribution maps for both linear and nonlinear datasets: 
\begin{enumerate}
    \item The first one, previously used in \cite{performative,stochasticperformative,outside,zeroinequality,pluginperformative}, affects only the feature $x_i$:
    \begin{equation}
        X\sim \mathcal{D}(\theta) = X_0 - d \theta,
    \label{linear_x_dw}
    \end{equation}
    where $X_0$ is the base data distribution, and $d$ is the parameter that reflects the sensitivity of data distribution map. We set $d = \left[30,40,50\right]$ for linear datasets and $d = \left[0.3,0.5,0.7\right]$ for nonlinear datasets. 

    The direct calculation of Eq.~(\ref{linear_x_dw}) is not applicable in the nonlinear methods (NP$^2$M$^2$ and RRM NN). For our method NP$^2$M$^2$, the nonlinear model $\theta = \sum_{i=1}^n \alpha_i y_i \phi(x_i)$ can not be calculated with the subtraction, as the $\phi(x_i)$ is in the high-dimensional space that can not be represented directly. For RRM NN, the parameters in neural networks are much more than the data dimensions, and the subtraction can not be performed. 
    Thus, we simulate the data distribution map given in Eq.~(\ref{linear_x_dw}). The physical meaning of this data distribution map is that all samples change their features, with the aim of being classified as negative by the classifier. Each sample randomly generates data points within a radius of $d$, and selects a point with the lowest classification score as its new representation after the data distribution shift. Specifically, the simulation of this data distribution map is performed as follows:
    \begin{enumerate}
        \item For each sample $x_i$, we generate $100$ random data points $x_{r\ i},r = 1,2,\cdots,100$. The distance of each random data points $x_{r\ i}$ from the original sample $x_i$ is smaller than or equal to $d$, where $d$ is the parameter that affects the sensitivity of data distribution map in Eq.~(\ref{linear_x_dw}).
        
        \item Calculate the function value $f(x_{r\ i})$ of all random data points $x_{r\ i},r = 1,2,\cdots,100$.
        
        \item Select the $x_{r^*\ i}$ that $x_{r^*\ i} = \arg \min_r f(x_{r\ i}),r = 1,2,\cdots,100$ as the new representation of this sample after data distribution shift.
    \end{enumerate}

    \item The second one affects only the label $y_i$:
    \begin{equation}
        y \sim \mathcal{D}(\theta) = \begin{cases}
            y_0,  & \text{ if } y f_\theta(x) < 0, \\
            y_0,  & \text{ if } y f_\theta(x) > 0 \text{ with probability } 1-p,\\
            -y_0,  & \text{ if } y f_\theta(x) > 0 \text{ with probability } p,\\
        \end{cases}
    \label{linear_y_dw}
    \end{equation}
    where $p = \frac{e^{d p^*}}{1+e^{d p^*}}$, with $p^* = \begin{cases} \frac{f_\theta(x_i)}{\max(f_\theta(x_i))} & \text{ if } y_i = +1 \\ \frac{f_\theta(x_i)}{\min(f_\theta(x_i))} & \text{ if } y_i = -1  \end{cases}$ indicating the relative position of data to the decision plane, where $f_\theta(x_i)$ denotes the prediction value of data $x_i$ by model $f$ with parameters $\theta$, $\max(f_\theta(x_i))$ and $\min(f_\theta(x_i))$ are the maximum and minimum prediction value over the dataset. In this data distribution map, smaller values of $p^*$ correspond to data closer to the decision plane and higher probability of label-changing. Similar to the first data distribution map, $d$ is the parameter that adjusts the performative effect, with small values leading to high probability of label-changing. In our experiments, we set $d = \left[5,7.5,10\right]$ for both linear and nonlinear datasets.
        
\end{enumerate}

\begin{table}[!t]
\centering
\caption{Performance comparison in terms of classification accuracy (ACC) for the proposed NP$^2$M$^2$ and six baseline approaches on linear synthetic dataset. Here, $\mathcal{D}(\theta)$ denotes the applied data distribution map, and $d$ denotes the parameter in the corresponding $\mathcal{D}(\theta)$. For each method under each setting, the average result and standard deviation over $10$ trials are reported. For each setting, the best result is highlighted in bold, and the second-best result is underlined.}
\vspace{-0.2cm}
\begin{adjustbox}{width=\textwidth,center=\textwidth}
{
\begin{tabular}{l|ccc|ccc}
\toprule
& \multicolumn{3}{c|}{$\mathcal{D}(\theta)$: Eq.~(\ref{linear_x_dw})} & \multicolumn{3}{c}{$\mathcal{D}(\theta)$: Eq.~(\ref{linear_y_dw})} \\
\cmidrule{2-7}      & $d = 30$ & $d = 40$ & $d = 50$ & $d = 5$ & $d = 7.5$ & $d = 10$ \\
\midrule
NP$^2$M$^2$ & 0\textbf{.838 $\pm$ 0.011} & \textbf{0.821 $\pm$ 0.019} & \textbf{0.800 $\pm$ 0.024} & \textbf{0.656 $\pm$ 0.013} & \underline{0.731 $\pm$ 0.009} & \underline{0.775 $\pm$ 0.008} \\
RRM LR & \underline{0.680 $\pm$ 0.037} & 0.537 $\pm$ 0.023 & 0.509 $\pm$ 0.009 & 0.601 $\pm$ 0.012 & 0.672 $\pm$ 0.011 & 0.717 $\pm$ 0.009 \\
RGD LR & 0.538 $\pm$ 0.011 & 0.514 $\pm$ 0.006 & 0.506 $\pm$ 0.004 & 0.603 $\pm$ 0.013 & 0.672 $\pm$ 0.010 & 0.717 $\pm$ 0.010 \\
RRM NN & 0.645 $\pm$ 0.032 & \underline{0.592 $\pm$ 0.031} & \underline{0.552 $\pm$ 0.023} & \underline{0.650 $\pm$ 0.014} & \textbf{0.744 $\pm$ 0.012} & \textbf{0.812 $\pm$ 0.013} \\
TSA   & 0.519 $\pm$ 0.005 & 0.507 $\pm$ 0.002 & 0.503 $\pm$ 0.001 & 0.598 $\pm$ 0.011 & 0.662 $\pm$ 0.010 & 0.705 $\pm$ 0.009 \\
PerfGD & 0.532 $\pm$ 0.010 & 0.513 $\pm$ 0.006 & 0.506 $\pm$ 0.003 & 0.601 $\pm$ 0.014 & 0.666 $\pm$ 0.010 & 0.712 $\pm$ 0.010 \\
DFO($\lambda$) & 0.509 $\pm$ 0.018 & 0.503 $\pm$ 0.004 & 0.501 $\pm$ 0.003 & 0.455 $\pm$ 0.073 & 0.479 $\pm$ 0.083 & 0.492 $\pm$ 0.091 \\
\bottomrule
\end{tabular}%
}
\end{adjustbox}
\label{synthetic_result_linear_acc}
\vspace{0.5cm}
\caption{Performance comparison in terms of model consistency (MC) for the proposed NP$^2$M$^2$ and six baseline approaches on linear synthetic dataset. Here, $\mathcal{D}(\theta)$ denotes the applied data distribution map, and $d$ denotes the parameter in the corresponding $\mathcal{D}(\theta)$. For each method under each setting, the average result and standard deviation over $10$ trials are reported. For each setting, the best result is highlighted in bold, and the second-best result is underlined.}
\vspace{-0.2cm}
\begin{adjustbox}{width=\textwidth,center=\textwidth}
{
\begin{tabular}{l|ccc|ccc}
\toprule
& \multicolumn{3}{c|}{$\mathcal{D}(\theta)$: Eq.~(\ref{linear_x_dw})} & \multicolumn{3}{c}{$\mathcal{D}(\theta)$: Eq.~(\ref{linear_y_dw})} \\
\cmidrule{2-7}      & $d = 30$ & $d = 40$ & $d = 50$ & $d = 5$ & $d = 7.5$ & $d = 10$ \\
\midrule
NP$^2$M$^2$ & \textbf{0.998 $\pm$ 0.000} & \textbf{0.998 $\pm$ 0.000} & \textbf{0.997 $\pm$ 0.001} & \textbf{0.620 $\pm$ 0.029} & \textbf{0.733 $\pm$ 0.026} & \textbf{0.789 $\pm$ 0.015} \\
RRM LR & 0.863 $\pm$ 0.061 & 0.763 $\pm$ 0.066 & 0.860 $\pm$ 0.049 & 0.445 $\pm$ 0.055 & \underline{0.624 $\pm$ 0.039} & \underline{0.707 $\pm$ 0.035} \\
RGD LR & 0.920 $\pm$ 0.010 & 0.916 $\pm$ 0.010 & 0.916 $\pm$ 0.010 & 0.441 $\pm$ 0.049 & 0.615 $\pm$ 0.034 & 0.702 $\pm$ 0.033 \\
RRM NN & 0.011 $\pm$ 0.034 & 0.004 $\pm$ 0.035 & 0.019 $\pm$ 0.039 & 0.022 $\pm$ 0.020 & 0.015 $\pm$ 0.026 & 0.014 $\pm$ 0.031 \\
TSA   & 0.915 $\pm$ 0.014 & 0.915 $\pm$ 0.014 & 0.917 $\pm$ 0.014 & 0.422 $\pm$ 0.053 & 0.576 $\pm$ 0.038 & 0.658 $\pm$ 0.034 \\
PerfGD & \underline{0.966 $\pm$ 0.007} & \underline{0.928 $\pm$ 0.010} & \underline{0.917 $\pm$ 0.010} & 0.445 $\pm$ 0.051 & 0.612 $\pm$ 0.036 & 0.692 $\pm$ 0.031 \\
DFO($\lambda$) & 0.462 $\pm$ 0.432 & 0.469 $\pm$ 0.421 & 0.466 $\pm$ 0.429 & \underline{0.473 $\pm$ 0.422} & 0.468 $\pm$ 0.425 & 0.455 $\pm$ 0.441 \\
\bottomrule
\end{tabular}%
}
\end{adjustbox}
\label{synthetic_result_linear_mc}
\end{table}%

\begin{table*}[!t]
\centering
\caption{Performance comparison in terms of classification accuracy (ACC) for the proposed NP$^2$M$^2$ and six baseline approaches on nonlinear synthetic dataset. Here, $\mathcal{D}(\theta)$ denotes the applied data distribution map, and $d$ denotes the parameter in the corresponding $\mathcal{D}(\theta)$. For each method under each setting, the average result and standard deviation over $10$ trials are reported. For each setting, the best result is highlighted in bold, and the second-best result is underlined.}
\vspace{-0.2cm}
\begin{adjustbox}{width=\textwidth,center=\textwidth}
{
\begin{tabular}{l|ccc|ccc}
\toprule
      & \multicolumn{3}{c|}{$\mathcal{D}(\theta)$: Eq.~(\ref{linear_x_dw})} & \multicolumn{3}{c}{$\mathcal{D}(\theta)$: Eq.~(\ref{linear_y_dw})} \\
\cmidrule{2-7}      & $d = 0.3$ & $d = 0.5$ & $d = 0.7$ & $d = 5$ & $d = 7.5$ & $d = 10$ \\
\midrule
NP$^2$M$^2$ & \textbf{0.982 $\pm$ 0.002} & \textbf{0.982 $\pm$ 0.001} & \textbf{0.982 $\pm$ 0.002} & \textbf{0.842 $\pm$ 0.006} & \textbf{0.897 $\pm$ 0.005} & \textbf{0.925 $\pm$ 0.004} \\
RRM LR & 0.389 $\pm$ 0.018 & 0.409 $\pm$ 0.015 & 0.432 $\pm$ 0.022 & 0.475 $\pm$ 0.024 & 0.488 $\pm$ 0.023 & 0.489 $\pm$ 0.022 \\
RGD LR & 0.387 $\pm$ 0.020 & 0.408 $\pm$ 0.017 & 0.437 $\pm$ 0.020 & 0.478 $\pm$ 0.012 & 0.488 $\pm$ 0.016 & 0.490 $\pm$ 0.023 \\
RRM NN & \underline{0.974 $\pm$ 0.006} & \underline{0.968 $\pm$ 0.012} & \underline{0.961 $\pm$ 0.011} & \underline{0.826 $\pm$ 0.010} & \underline{0.879 $\pm$ 0.009} & \underline{0.911 $\pm$ 0.008} \\
TSA & 0.340 $\pm$ 0.003 & 0.340 $\pm$ 0.002 & 0.378 $\pm$ 0.002 & 0.375 $\pm$ 0.006 & 0.401 $\pm$ 0.005 & 0.416 $\pm$ 0.005 \\
PerfGD & 0.340 $\pm$ 0.003 & 0.340 $\pm$ 0.003 & 0.378 $\pm$ 0.002 & 0.374 $\pm$ 0.005 & 0.400 $\pm$ 0.005 & 0.418 $\pm$ 0.005 \\
DFO($\lambda$) & 0.490 $\pm$ 0.036 & 0.476 $\pm$ 0.039 & 0.475 $\pm$ 0.031 & 0.425 $\pm$ 0.032 & 0.466 $\pm$ 0.040 & 0.476 $\pm$ 0.035 \\
\bottomrule
\end{tabular}%
}
\end{adjustbox}
\label{synthetic_result_nonlinear_acc}
\vspace{0.5cm}
\caption{Performance comparison in terms of model consistency (MC) for the proposed NP$^2$M$^2$ and six baseline approaches on nonlinear synthetic dataset. Here, $\mathcal{D}(\theta)$ denotes the applied data distribution map, and $d$ denotes the parameter in the corresponding $\mathcal{D}(\theta)$. For each method under each setting, the average result and standard deviation over $10$ trials are reported. For each setting, the best result is highlighted in bold, and the second-best result is underlined.}
\vspace{-0.2cm}
\begin{adjustbox}{width=\textwidth,center=\textwidth}
{
\begin{tabular}{l|ccc|ccc}
\toprule
& \multicolumn{3}{c|}{$\mathcal{D}(\theta)$: Eq.~(\ref{linear_x_dw})} & \multicolumn{3}{c}{$\mathcal{D}(\theta)$: Eq.~(\ref{linear_y_dw})} \\
\cmidrule{2-7}      & $d = 0.3$ & $d = 0.5$ & $d = 0.7$ & $d = 5$ & $d = 7.5$ & $d = 10$ \\
\midrule
NP$^2$M$^2$ & \textbf{0.991 $\pm$ 0.001} & \textbf{0.993 $\pm$ 0.001} & \textbf{0.992 $\pm$ 0.001} & \textbf{0.985 $\pm$ 0.002} & \textbf{0.986 $\pm$ 0.002} & \textbf{0.986 $\pm$ 0.002} \\
RRM LR & 0.023 $\pm$ 0.215 & -0.013 $\pm$ 0.181 & -0.008 $\pm$ 0.164 & -0.371 $\pm$ 0.218 & -0.230 $\pm$ 0.134 & -0.123 $\pm$ 0.127 \\
RGD LR & -0.006 $\pm$ 0.199 & -0.004 $\pm$ 0.194 & -0.026 $\pm$ 0.176 & -0.369 $\pm$ 0.093 & -0.230 $\pm$ 0.074 & -0.125 $\pm$ 0.112 \\
RRM NN & 0.099 $\pm$ 0.066 & 0.117 $\pm$ 0.062 & 0.135 $\pm$ 0.063 & 0.071 $\pm$ 0.061 & 0.101 $\pm$ 0.061 & 0.082 $\pm$ 0.055 \\
TSA   & -0.002 $\pm$ 0.219 & 0.011 $\pm$ 0.213 & 0.011 $\pm$ 0.227 & 0.007 $\pm$ 0.207 & 0.015 $\pm$ 0.227 & 0.021 $\pm$ 0.234 \\
PerfGD & -0.044 $\pm$ 0.216 & -0.006 $\pm$ 0.231 & -0.004 $\pm$ 0.210 & -0.947 $\pm$ 0.021 & -0.894 $\pm$ 0.045 & -0.805 $\pm$ 0.077 \\
DFO($\lambda$) & \underline{0.444 $\pm$ 0.434} & \underline{0.424 $\pm$ 0.440} & \underline{0.399 $\pm$ 0.464} & \underline{0.414 $\pm$ 0.463} & \underline{0.440 $\pm$ 0.412} & \underline{0.425 $\pm$ 0.423} \\
\bottomrule
\end{tabular}%
}
\end{adjustbox}
\label{synthetic_result_nonlinear_mc}
\end{table*}%

\subsubsection{Result Analysis}

Tables~\ref{synthetic_result_linear_acc}~and~\ref{synthetic_result_linear_mc} present the accuracy and model consistency, respectively, of all methods on the linear synthetic dataset. Our method, NP$^2$M$^2$, achieves the highest accuracy in 4 out of 6 settings and ranks the second in the remaining two, as shown in Tables~\ref{synthetic_result_linear_acc}. For model consistency, NP$^2$M$^2$ outperforms all baselines across all settings, as detailed in Tables~\ref{synthetic_result_linear_mc}. In the two settings where RRM NN achieves the highest accuracy and NP$^2$M$^2$ ranks second, the data distribution map $\mathcal{D}(\theta)$ in Eq.~(\ref{linear_y_dw}) is less sensitive ($d = 7.5, 10$). Conversely, when $\mathcal{D}(\theta)$ is more sensitive ($d = 5$), NP$^2$M$^2$ attains the highest accuracy by actively adapting to the sensitive distribution map through appropriate model deployment, ensuring performative stability throughout the iterations.


Tables~\ref{synthetic_result_nonlinear_acc}~and~\ref{synthetic_result_nonlinear_mc} present the accuracy and model consistency, respectively, of all methods on the nonlinear synthetic dataset. Our method consistently ranks first across all settings in terms of both accuracy and model consistency, demonstrating its superior performance on the complex nonlinear dataset. Notably, NP$^2$M$^2$ achieves model consistency values above $0.98$, compared to the best baseline's maximum of $0.444$, highlighting the exceptional model stability of our method. 
The strong performance of our method is due to the deployment of an appropriate loss function based on the measured sensitivity of $\mathcal{D}(\theta)$. This adaptability allows it to perform well across various datasets. In contrast, existing methods, despite their theoretical guarantees, rely on idealized conditions such as convexity of performative loss, bounded gradient of performative loss, and $\beta$-smoothness. 
These conditions are often difficult to satisfy in diverse scenarios, leading to discrepancies between theoretical guarantees and empirical results.

\subsection{Company Bankruptcy Prediction}
In this subsection, we evaluate our method using the Taiwanese Bankruptcy Prediction dataset \cite{bankruptcyprediction} with a data distribution map $\mathcal{D}(\theta)$ derived from a real-world design.

\subsubsection{Dataset}
This dataset contains financial analysis data for $6,819$ companies in Taiwan, including $220$ bankrupt companies ($y = +1$) and $6,599$ solvent companies ($y = -1$). To address the label imbalance, we apply the NearMiss-3 approach~\cite{nearmiss,soltanzadeh2023addressing} to under-sample the solvent companies to $220$. Given the nonlinear nature of the dataset, we employ the RBF kernel in our method: $K(x,x^\prime) = e^{-\frac{\|x-x^\prime\|}{2\sigma^2}}$, which yielded the best overall performance compared with other kernels. We set the kernel parameter $\sigma = 0.1$, determined by cross-validation.

\subsubsection{Data Distribution Map}
All companies adjust their features in response to prediction results, aiming to avoid bankruptcy or improve their financial status. 
In this dataset, $85$ features are performative (modifiable by companies), and $10$ features are non-performative (unchangeable). Specifically, the data distribution map is designed as follows:
\begin{itemize}
    \item For all performative features $ X_{P}$, we have:
    \begin{equation}
        X_{P} \sim \mathcal{D}(\theta) = X_{0\  P} - d \theta,
    \end{equation}
    where $X_{0\  P}$ represents performative features in the base data.
    
    \item For labels $y$, only those data with $y = +1$ (bankruptcy) can change:
    \begin{equation}
        y \sim \mathcal{D}(\theta) = \begin{cases}
            y_0,  & \text{ with probability } 1-p,\\
            -y_0, & \text{ with probability } p,\\
        \end{cases} \ \forall y_0 = +1,
        \label{d_thata_company_y}
    \end{equation}
    where $p = \frac{e^{b p^*}}{1+e^{b p^*}}$, with $p^* = \frac{f_\theta(x_i)}{\max(f_\theta(x_i))}$ indicating the relative position of data to the decision plane.
\end{itemize}
The parameters are set as $d = \left[100,250,500,750,1000\right]$ and $b = -0.01 d + 15$ (i.e., $b = \left[14,12.5,10,7.5,5\right]$). With this setting, larger $d$ values lead to greater shifts in features $X$ and higher probabilities $p$ of the change of labels $y$, that is, in essence, more sensitive data distribution maps $\mathcal{D}(\theta)$. 

\begin{table*}[!t]
\centering
\caption{Performance comparison in terms of classification accuracy (ACC) for the proposed NP$^2$M$^2$ and six baseline approaches on the Company Bankruptcy Prediction dataset. Here, $d$ denotes the parameter in $\mathcal{D}(\theta)$. For each method under each setting, the average result and standard deviation over $10$ trials are reported. For each setting, the best result is highlighted in bold, and the second-best result is underlined.}
\vspace{-0.2cm}
\begin{adjustbox}{width=\textwidth,center=\textwidth}
{
\begin{tabular}{l|ccccc}
\toprule
& $d = 100$ & $d = 250$ & $d = 500$ & $d = 750$ & $d = 1000$ \\
\midrule
NP$^2$M$^2$ & \textbf{0.877 $\pm$ 0.004} & \textbf{0.873 $\pm$ 0.004} & \textbf{0.862 $\pm$ 0.004} & \textbf{0.843 $\pm$ 0.006} & \textbf{0.813 $\pm$ 0.007} \\
RRM LR & 0.783 $\pm$ 0.011 & 0.766 $\pm$ 0.013 & 0.747 $\pm$ 0.011 & 0.736 $\pm$ 0.010 & 0.730 $\pm$ 0.011 \\
RGD LR & 0.797 $\pm$ 0.008 & 0.786 $\pm$ 0.009 & 0.771 $\pm$ 0.008 & 0.758 $\pm$ 0.008 & \underline{0.744 $\pm$ 0.008} \\
RRM NN & 0.740 $\pm$ 0.010 & 0.733 $\pm$ 0.011 & 0.726 $\pm$ 0.010 & 0.723 $\pm$ 0.011 & 0.722 $\pm$ 0.009 \\
TSA & 0.713 $\pm$ 0.011 & 0.713 $\pm$ 0.011 & 0.712 $\pm$ 0.013 & 0.707 $\pm$ 0.021 & 0.686 $\pm$ 0.029 \\
PerfGD & \underline{0.804 $\pm$ 0.009} & \underline{0.794 $\pm$ 0.009} & \underline{0.775 $\pm$ 0.009} & \underline{0.759 $\pm$ 0.009} & 0.735 $\pm$ 0.012 \\
DFO($\lambda$) & 0.536 $\pm$ 0.036 & 0.534 $\pm$ 0.041 & 0.531 $\pm$ 0.049 & 0.537 $\pm$ 0.046 & 0.546 $\pm$ 0.042 \\
\bottomrule
\end{tabular}%
}
\end{adjustbox}
\label{company_result_acc}
\vspace{0.5cm}
\caption{Performance comparison in terms of model consistency (MC) for the proposed NP$^2$M$^2$ and six baseline approaches on the Company Bankruptcy Prediction dataset. Here, $d$ denotes the parameter in $\mathcal{D}(\theta)$. For each method under each setting, the average result and standard deviation over $10$ trials are reported. For each setting, the best result is highlighted in bold, and the second-best result is underlined.}
\vspace{-0.2cm}
\begin{adjustbox}{width=\textwidth,center=\textwidth}
{
\begin{tabular}{l|ccccc}
\toprule
      & $d = 100$ & $d = 250$ & $d = 500$ & $d = 750$ & $d = 1000$ \\
\midrule
NP$^2$M$^2$ & \textbf{0.792 $\pm$ 0.012} & \textbf{0.753 $\pm$ 0.013} & \textbf{0.693 $\pm$ 0.014} & \textbf{0.614 $\pm$ 0.015} & \textbf{0.514 $\pm$ 0.018} \\
RRM LR & 0.634 $\pm$ 0.026 & 0.587 $\pm$ 0.031 & 0.531 $\pm$ 0.035 & 0.486 $\pm$ 0.034 & 0.441 $\pm$ 0.035 \\
RGD LR & 0.669 $\pm$ 0.027 & 0.639 $\pm$ 0.027 & 0.580 $\pm$ 0.030 & 0.526 $\pm$ 0.033 & 0.461 $\pm$ 0.031 \\
RRM NN & 0.019 $\pm$ 0.020 & 0.017 $\pm$ 0.020 & 0.016 $\pm$ 0.019 & 0.015 $\pm$ 0.019 & 0.018 $\pm$ 0.020 \\
TSA   & 0.152 $\pm$ 0.041 & 0.133 $\pm$ 0.037 & 0.139 $\pm$ 0.041 & 0.113 $\pm$ 0.038 & 0.106 $\pm$ 0.036 \\
PerfGD & \underline{0.673 $\pm$ 0.025} & \underline{0.644 $\pm$ 0.021} & \underline{0.594 $\pm$ 0.030} & \underline{0.528 $\pm$ 0.030} & 0.467 $\pm$ 0.033 \\
DFO($\lambda$) & 0.471 $\pm$ 0.422 & 0.470 $\pm$ 0.423 & 0.470 $\pm$ 0.424 & 0.467 $\pm$ 0.427 & \underline{0.469 $\pm$ 0.423} \\
\bottomrule
\end{tabular}%
}
\end{adjustbox}
\label{company_result_mc}
\end{table*}%

\subsubsection{Result Analysis}
Tables \ref{company_result_acc}~and~\ref{company_result_mc} report the accuracy and model consistency, respectively, of all methods on the Company Bankruptcy Prediction dataset. 
In this nonlinear dataset, performative effects influence both features and labels, presenting a more complex scenario than in synthetic datasets. 
Existing methods, based on rigorous, uncontrollable conditions, underestimate the complexity of real-world applications. For instance, DFO($\lambda$) and PerfGD explore the performative loss gradient during iterations. However, in this $\mathcal{D}(\theta)$, label shifts cause discontinuous performative loss, preventing these methods from achieving optimal results. In contrast, the proposed method performs consistently well. It ranks the highest among all methods in both accuracy and model consistency across all settings, which is consistent with the results on synthetic datasets.

\begin{figure}[!t]
\vspace{-0.3cm}
\centering
    \subfloat[Accuracy with $d = 100$]{\includegraphics[width=0.5\linewidth]{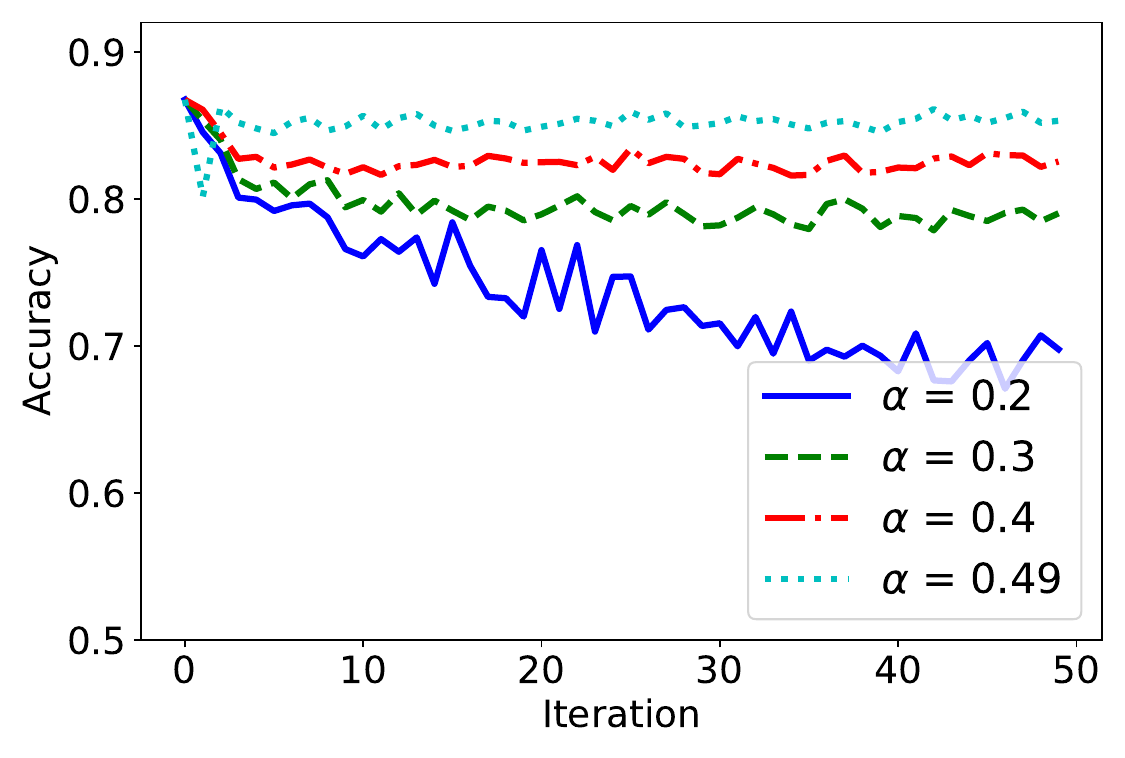}}
    \subfloat[Model Consistency with $d = 100$]{\includegraphics[width=0.5\linewidth]{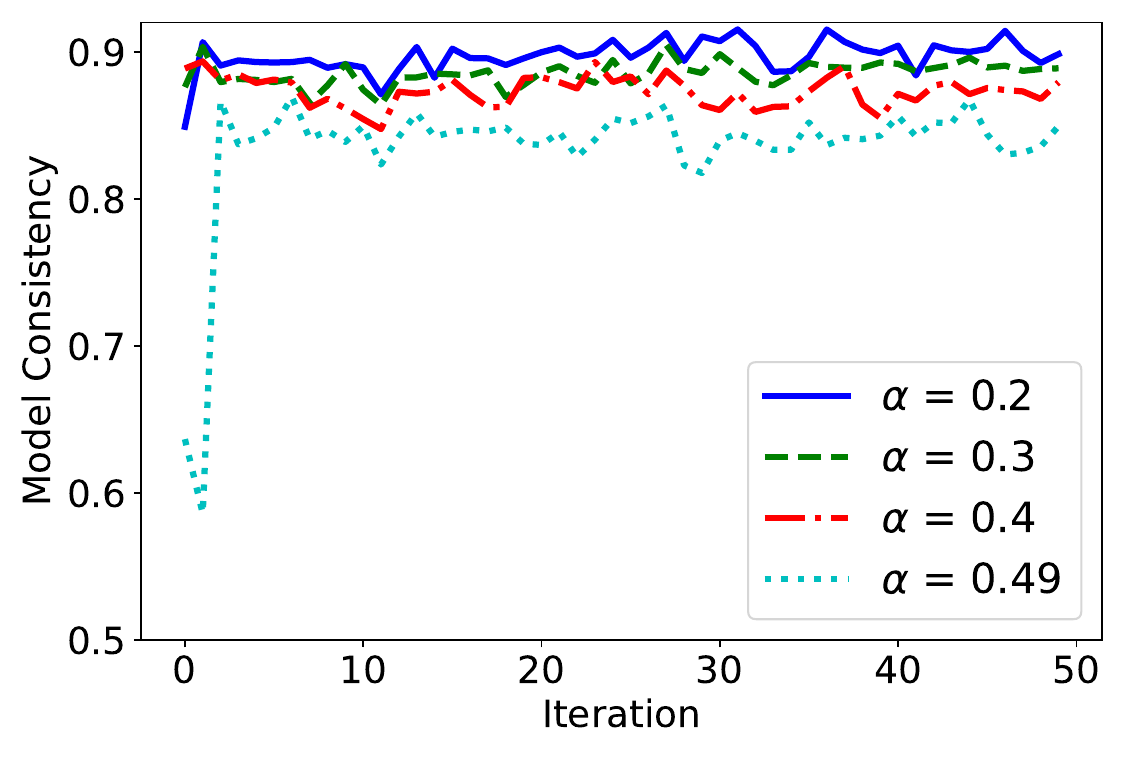}}\\
    \subfloat[Accuracy with $d = 250$]{\includegraphics[width=0.5\linewidth]{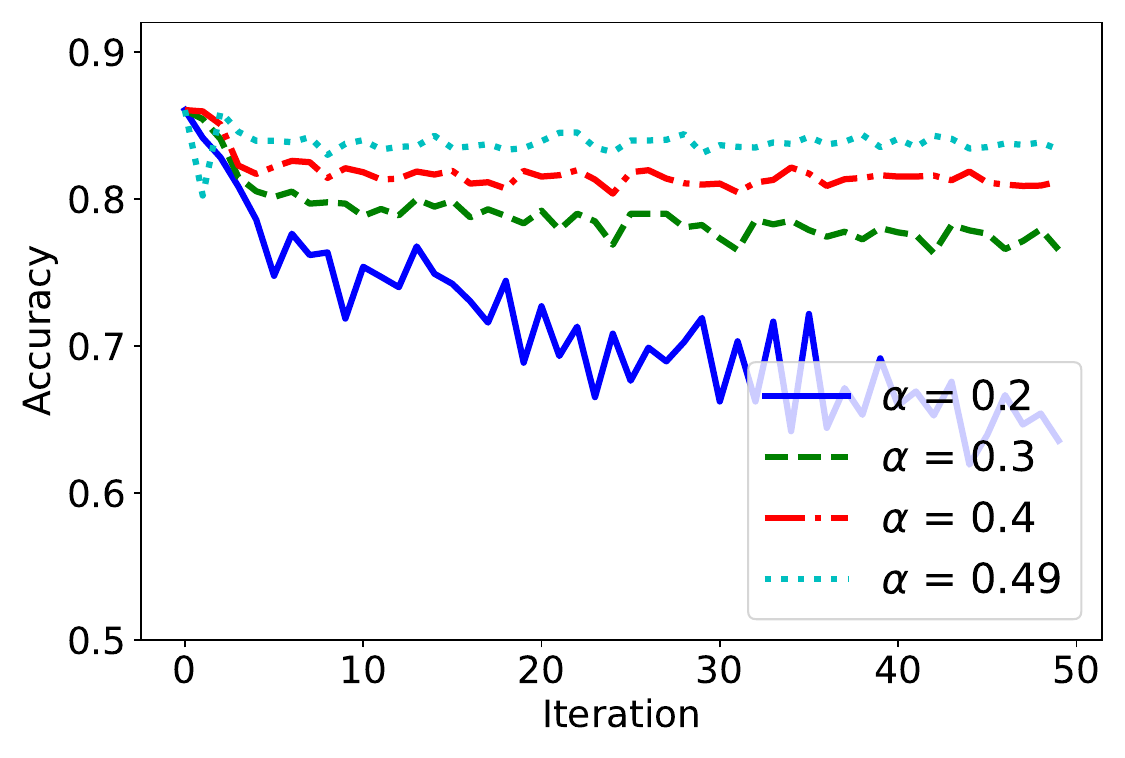}}
    \subfloat[Model Consistency with $d = 250$]{\includegraphics[width=0.5\linewidth]{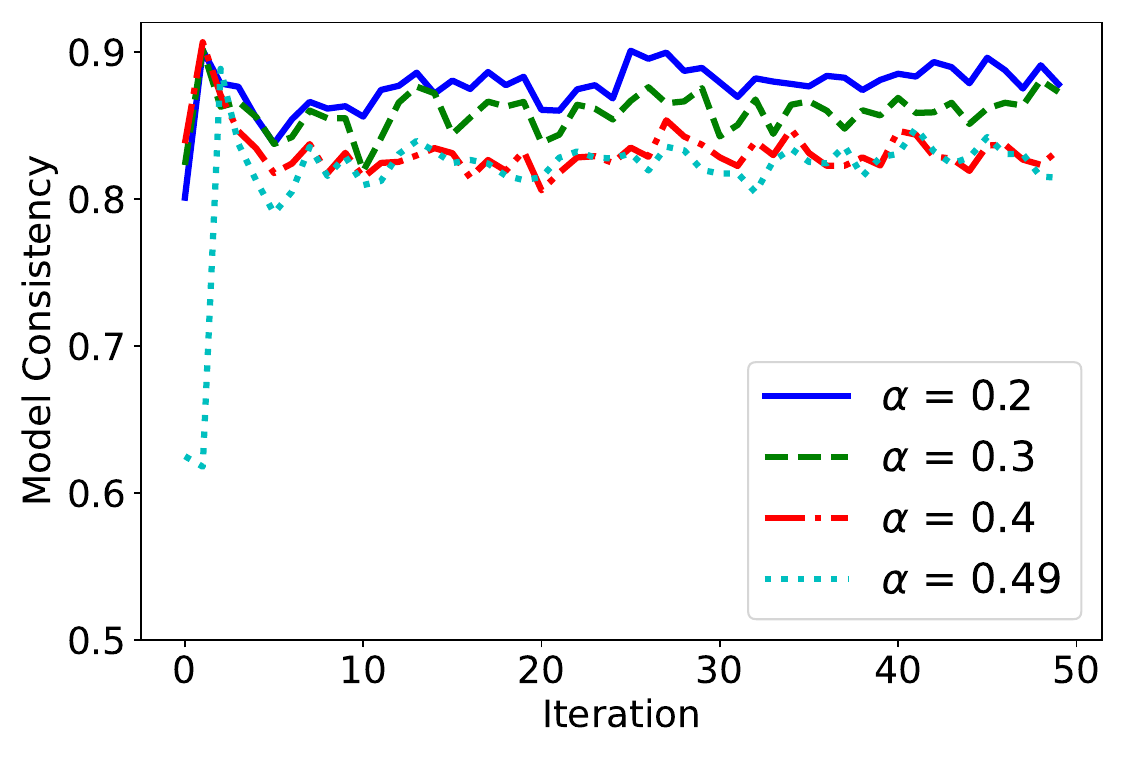}}\\
    \subfloat[Accuracy with $d = 500$]{\includegraphics[width=0.5\linewidth]{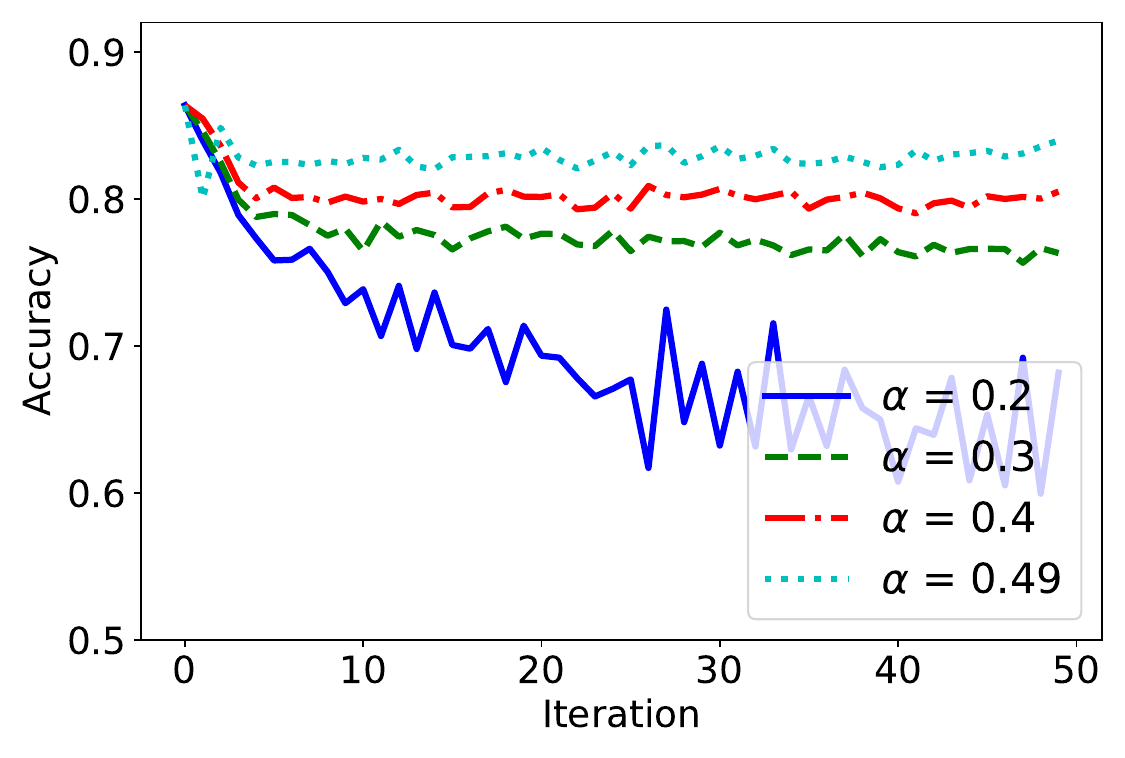}}
    \subfloat[Model Consistency with $d = 500$]{\includegraphics[width=0.5\linewidth]{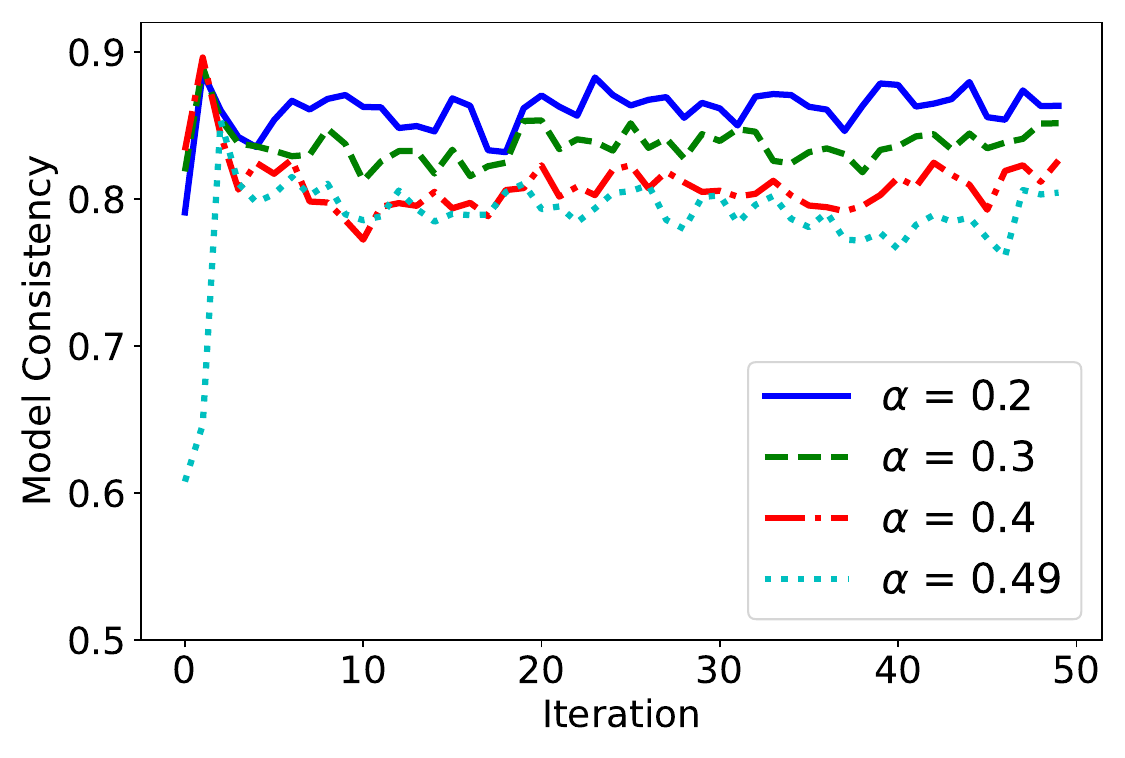}}
    \vspace{-0.2cm}
    \caption{Average accuracy and Model Consistency of NP$^2$M$^2$ with different $\alpha$ in Company Bankruptcy Prediction dataset with $d = 100,250,500,750,1000$ over 10 trials}
    \vspace{0.5cm}
    \label{bank_1_a}
\end{figure}

\begin{figure}[!t]
\ContinuedFloat
\centering
    \subfloat[Accuracy with $d = 750$]{\includegraphics[width=0.5\linewidth]{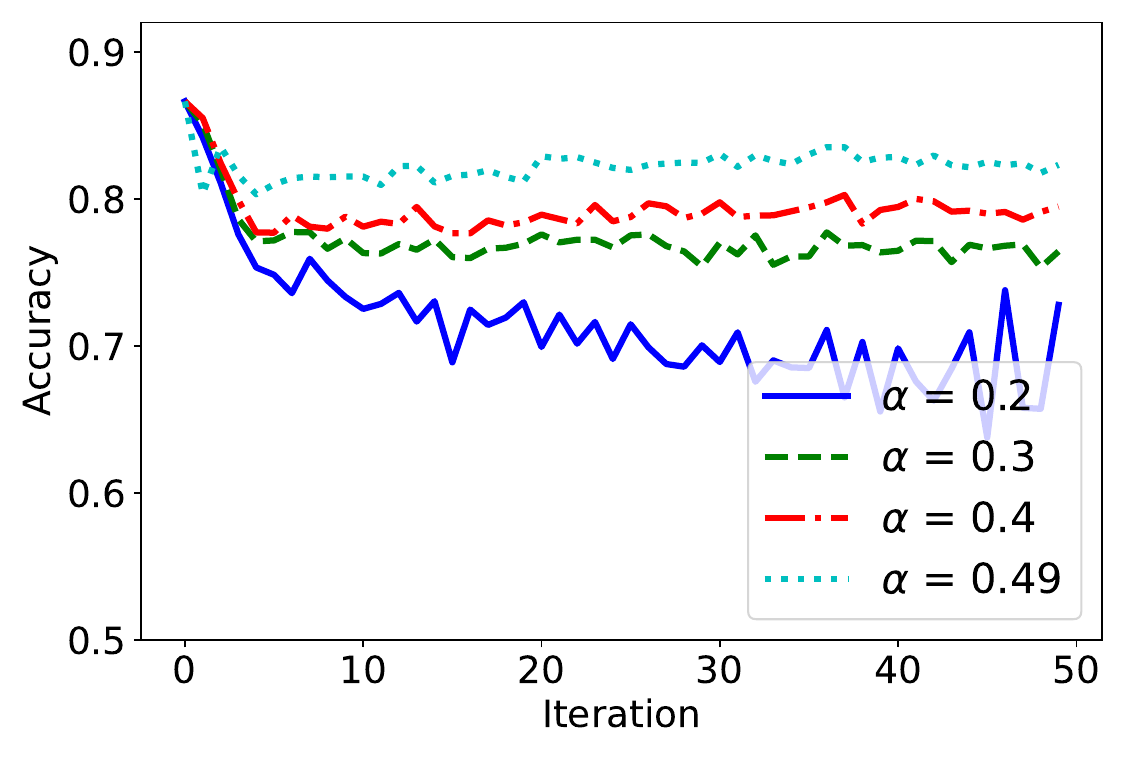}}
    \subfloat[Model Consistency with $d = 750$]{\includegraphics[width=0.5\linewidth]{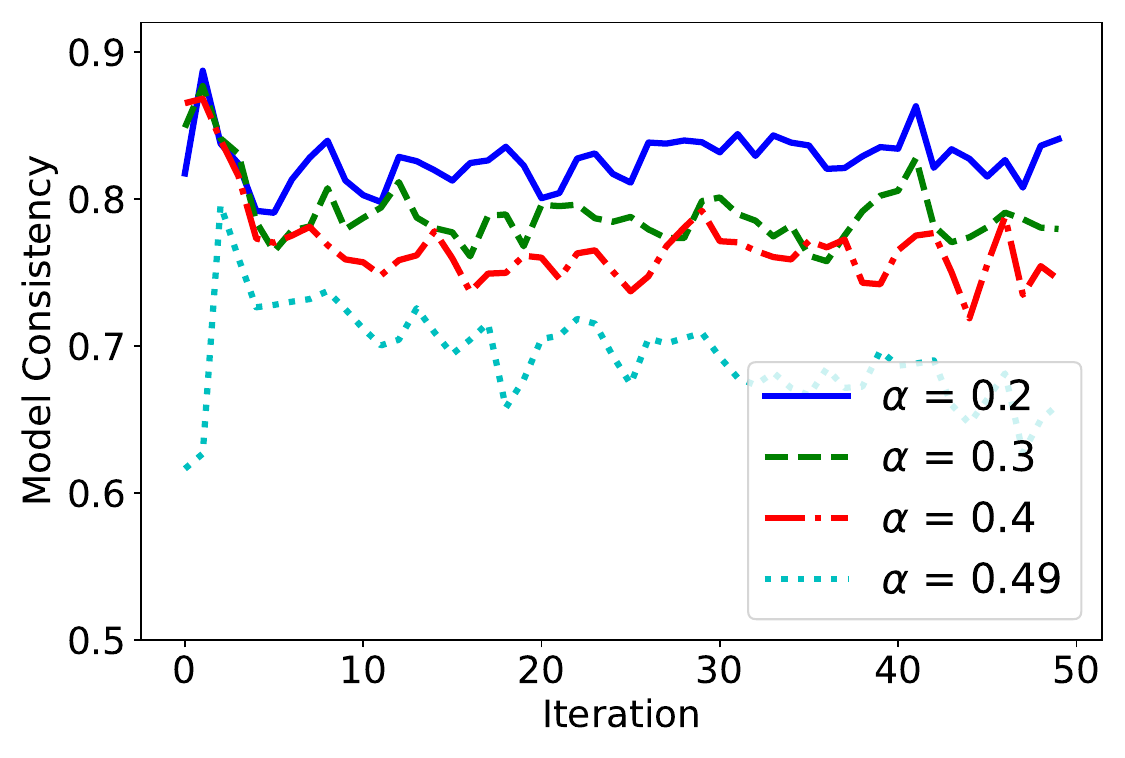}}\\
    \subfloat[Accuracy with $d = 1000$]{\includegraphics[width=0.5\linewidth]{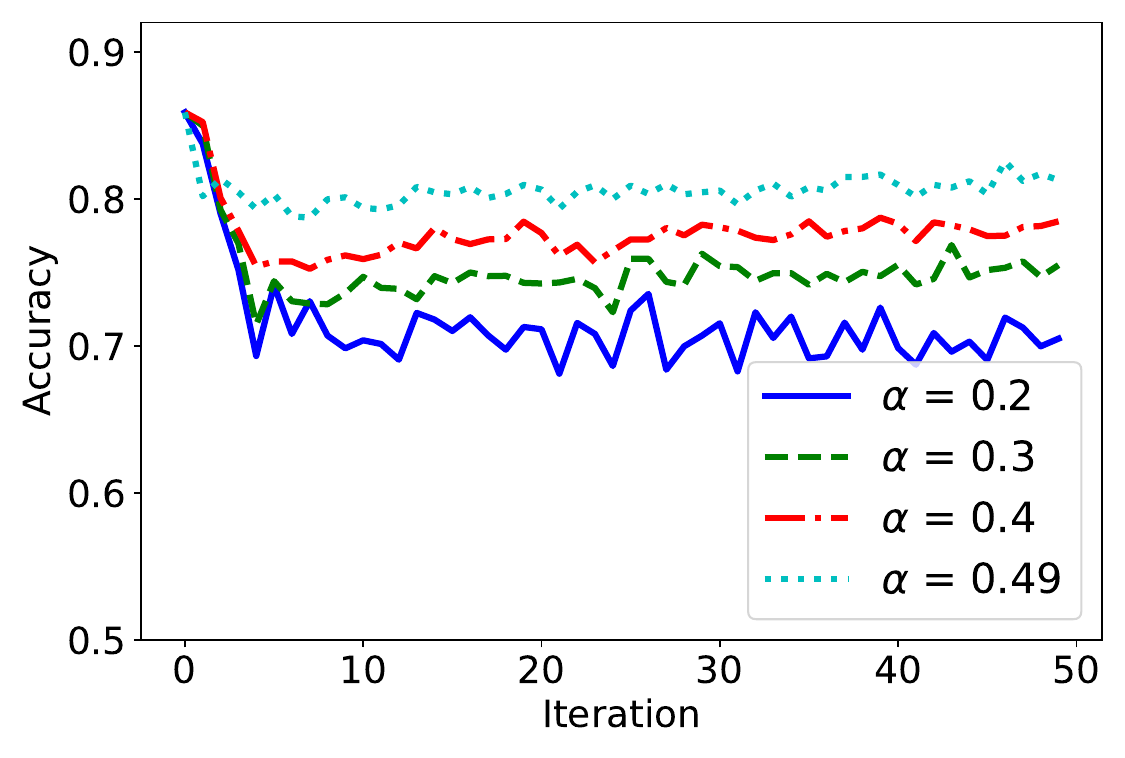}}
    \subfloat[Model Consistency with $d = 1000$]{\includegraphics[width=0.5\linewidth]{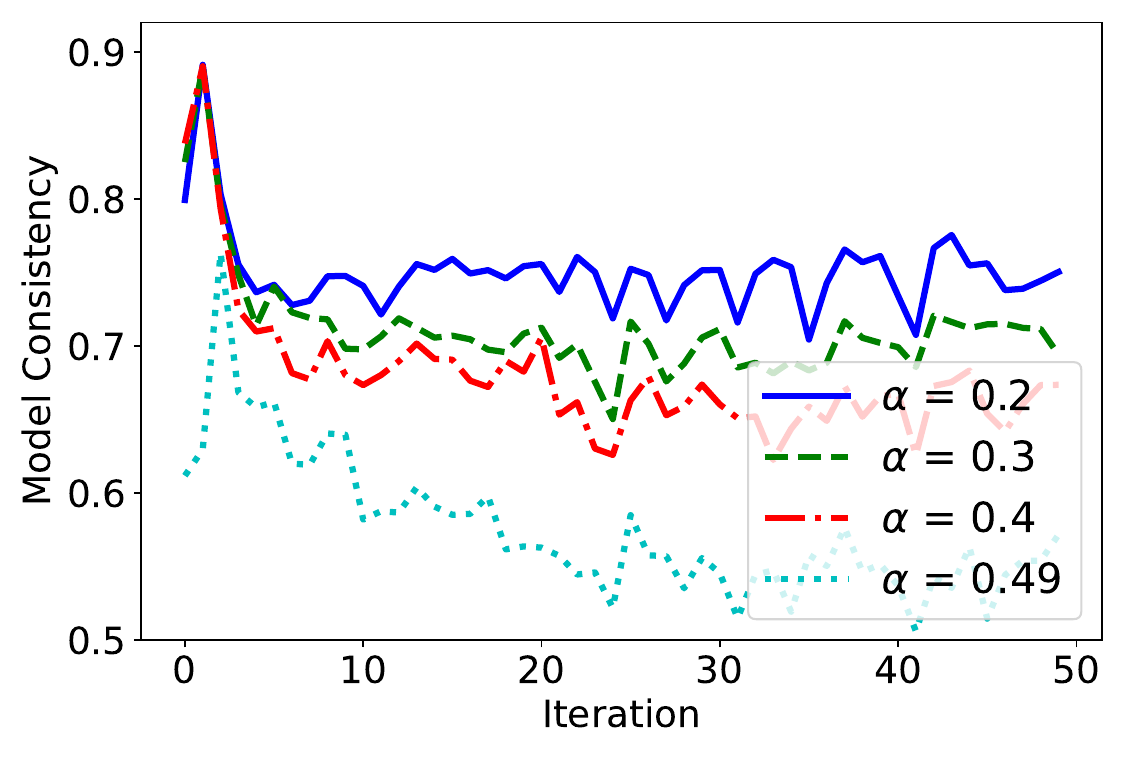}}
    \vspace{-0.2cm}
    \caption{Continued.}
\end{figure}

\subsection{Parameter Sensitivity Analysis} \label{ParameterSensitivity}

In this subsection, we conduct a parameter sensitivity analysis to evaluate the robustness of the proposed method with respect to parameter $\alpha$, which controls the convergence rate to performative stability. We evaluate the accuracy and model consistency on the Company Bankruptcy Prediction dataset with $\alpha = 0.2,0.3,0.4,0.49$. The average results over 10 trials with each $d$ ($d = 100$, $250$, $500$, $750$, $1000$) are shown in Fig.~\ref{bank_1_a}. Our method exhibits consistent performance across all values of $\alpha$, indicating its robustness to this key parameter. As established in Theorem~\ref{theorem2}, performative stability is guaranteed when $\varepsilon C < \frac{1}{2}$. Thus, for any $\alpha \in (0,0.5)$ in NP$^2$M$^2$, we ensure the performative stability with high probability. However, the accuracy for $\alpha = 0.2$ is slightly lower compared to other settings. The reason is that a smaller $\alpha$ results in a smaller $C$ in the objective function formulated in Eq.~(\ref{svmprime}), thus leading to suboptimal performance. Based on these findings, we set $\alpha = 0.49$ in all previous experiments, as it ensures the performative stability while maintaining accuracy. 

\section{Conclusion} \label{conclusion}
In this paper, we developed a novel learning framework for Nonlinear Performative Prediction via Maximum Margin approach (NP$^2$M$^2$). By relaxing out-of-control assumptions in existing work, we derived conditions for performative stability that are both theoretically useful and practically applicable. Based on our theoretical results, we introduced an algorithm that ensures performative stability of the predictive model. Performance comparison with state-of-the-art approaches demonstrates the effectiveness of our method. For future work, we plan to extend our work to the stateful setting \cite{performativestateful,li2022state}, where data distribution is determined not only by the deployed model but also by historical data distributions, presenting additional challenges for theoretical analysis and model design.


\section*{CRediT authorship contribution statement}

\textbf{Guangzheng Zhong:} Conceptualization, Formal analysis, Investigation, Methodology, Software, Writing – original draft
\textbf{Yang Liu:} Conceptualization, Formal analysis, Investigation, Methodology, Project administration, Writing – original draft, Writing – review and editing
\textbf{Jiming Liu:} Conceptualization, Investigation, Project administration, Writing – review and editing

\section*{Data availability}
The code and dataset are available at: `github.com/zhong-gz/NP2M2'

\section*{Declaration of competing interest}
The authors declare that they have no known competing financial interests or personal relationships that could have appeared to influence the work reported in this paper. 

\section*{Acknowledgements} 
This work was supported in part by the General Research Fund from the Research Grant Council of Hong Kong SAR under Projects RGC/HKBU12203122 and RGC/HKBU12200124, the NSFC/RGC Joint Research Scheme under Project N HKBU222/22, and the Guangdong Basic and Applied Basic Research Foundation under Project 2024A1515011837.

\bibliographystyle{elsarticle-num} 
\bibliography{mybib.bib}

\end{document}